**Scalable deep fusion of spaceborne lidar and synthetic aperture radar for global forest structural complexity mapping**

Tiago de Conto[1], John Armston[1], Ralph Dubayah[1]

[1]University of Maryland, 2181 Samuel J. LeFrak Hall, 7251 Preinkert Drive, College Park, MD 20742, United States of America.

Email: tiagodc@umd.edu

**Abstract**

Forest structural complexity metrics integrate multiple canopy attributes into a single value that reflects habitat quality and ecosystem function. Spaceborne lidar from the Global Ecosystem Dynamics Investigation (GEDI) has enabled mapping of structural complexity in temperate and tropical forests, but its sparse sampling limits continuous high-resolution mapping. We present a scalable, deep learning framework fusing GEDI observations with multimodal Synthetic Aperture Radar (SAR) datasets to produce global, high-resolution (25 m) wall-to-wall maps of forest structural complexity. Our adapted EfficientNetV2 architecture, trained on over 130 million GEDI footprints, achieves high performance (global $R^2 = 0.82$) with fewer than 400,000 parameters, making it an accessible tool that enables researchers to process datasets at any scale without requiring specialized computing infrastructure. The model produces accurate predictions with calibrated uncertainty estimates across biomes and time periods, preserving fine-scale spatial patterns. It has been used to generate a global, multi-temporal dataset of forest structural complexity from 2015 to 2022. Through transfer learning, this framework can be extended to predict additional forest structural variables with minimal computational cost. This approach supports continuous, multi-temporal monitoring of global forest structural dynamics and provides tools for biodiversity conservation and ecosystem management efforts in a changing climate.

## 1. Introduction

Forest structural complexity has emerged as a critical research topic at the intersection of remote sensing and forest ecology [1–3]. Various indices have been developed to estimate canopy complexity and although these metrics differ in their mathematical formulations and ecological foundations, they share a common definition: the three-dimensional occupancy and variability of tree structures [4–6]. This concept extends beyond habitat quality assessments to serve as a robust indicator of ecosystem health, function, and biodiversity [7].

Complexity indices often serve as abstract integrators, combining multiple structural metrics into a single value that reflects overall forest structural complexity [8–10]. Modern approaches to measuring forest structural complexity have relied on either image-based remote sensing or lidar point clouds [11]. Image-based remote sensing, while widely available, is limited in its ability to penetrate canopies and retrieve three-dimensional information. Conversely, airborne and terrestrial lidar systems provide highly detailed 3D structural measurements but are typically cost-prohibitive and impractical for large area applications.

To address these limitations, the Global Ecosystem Dynamics Investigation (GEDI) mission was launched in December 2018 as a spaceborne lidar system installed on the International Space Station [12]. GEDI employs full-waveform lidar technology, capturing discrete data points (footprints) sampled across the Earth's surface that provide direct measurements of forest vertical structure. This technology penetrates through gaps in canopy layers to detect understory components, capturing the vertical distribution of plant material throughout the entire canopy profile. Designed to observe the world's temperate and tropical forests (between 51.6° N and 51.6° S), GEDI has collected billions of observations across diverse biomes, providing unprecedented data on vegetation structure worldwide.

Leveraging these extensive waveform measurements from GEDI, we developed the Waveform Structural Complexity Index (WSCI) [13]. The WSCI integrates information from multiple structural attributes within GEDI footprints while establishing robust links with high-resolution 3D lidar point clouds. The WSCI gives global-scale estimation of forest complexity by capturing both canopy and understory structure—capabilities not previously attainable at this resolution or extent. However, because GEDI samples only a small fraction of the land surface, maps based on its sparse data must be aggregated to resolutions of 1 × 1 km or coarser [12]. Characterizing an area's structural complexity ideally requires contiguous footprints to account for inherent spatial and data variability. For many ecological and conservation applications such as habitat modeling and forest change monitoring, continuous high-resolution spatial information is essential, a capability that GEDI's sampling alone cannot provide.

While terrestrial or airborne lidar systems are ideal for creating continuous, high-resolution maps of forest structural complexity over time, deploying these technologies at landscape to global scales remains prohibitively expensive and logistically challenging. The quantity and spatial distribution of GEDI's sampling at a global scale far exceeds the capabilities of terrestrial and airborne lidar systems but has only been available since 2019. To address GEDI's spatial and temporal mapping limitations and reveal fine-scale patterns and gradients of structural complexity, we must develop methods to map continuously across landscapes by filling the gaps between GEDI footprints through integration with complementary remote sensing data sources [14,15].

To meet this challenge, the remote sensing community has increasingly used spaceborne imagery from both passive optical sensors and active Synthetic Aperture Radar (SAR) systems. SAR offers particular advantages for forest structure analysis because its signal responds directly to the electrical and geometric properties of vegetation elements [16,17]. SAR systems interact with forest structure in a wavelength-dependent manner: shorter wavelengths primarily interact with upper canopy elements, while longer wavelengths penetrate deeper into the forest canopy [18]. This wavelength-dependent interaction with forest structural elements establishes a functional complementarity between SAR and lidar measurements, as both technologies capture three-dimensional structural information. SAR systems provide the continuous spatial coverage absent in GEDI's sampling design, making SAR data particularly suitable for spatial data fusion approaches that can leverage the strengths of both remote sensing modalities.

Data fusion techniques harness information from multiple sources to derive improved estimates when a single mode is insufficient. In remote sensing, these methods typically support geographical extrapolation by combining sparse, spatially constrained data with widely available imagery [19,20]. Previous studies have demonstrated the complementarity between GEDI and SAR datasets through fusion techniques to extrapolate estimates of canopy height [14,15,21–24] and aboveground biomass [15,25–27]. While various approaches exist for fusing lidar and SAR data, deep learning has emerged as the leading approach for data fusion in remote sensing due to its flexibility in using spatial context information while integrating multiple inputs and learning complex non-linear relationships in the data [19,28].

The large volume of GEDI observations provides an unprecedented resource for training deep learning models for Earth observation. However, current deep learning architectures used in remote sensing are frequently computationally intensive, making them inaccessible to researchers with limited computing resources and challenging to deploy for global-scale, multi-temporal analyses. While these complex models do offer valuable capabilities such as the ability to estimate predictive uncertainty and provide essential information about prediction reliability [14,15,21], their computational demands limit their practical application. This creates a need for simpler, more efficient models that can still capture predictive uncertainty while remaining accessible for widespread use. Moreover, available pre-trained deep learning models are typically spatially or temporally constrained in extent or resolution, and none have specifically focused on predicting forest structural complexity at high spatial resolution through time.

Here, our goal is to expand the application domain of deep learning models for extrapolating GEDI observations by addressing four key objectives: (1) developing and validating a computationally efficient deep neural network that estimates GEDI WSCI and predicts its uncertainty wall-to-wall from multi-source spaceborne SAR datasets at 25 m resolution; (2) investigating the spatiotemporal relative importance of different input data sources; (3) evaluating the model's ability to describe local gradients of complexity at high resolution; and (4) understanding its potential as a foundational model for predicting other GEDI variables through transfer learning. We focus on WSCI as our target variable because it serves as an integrative metric highly correlated with many forest structural attributes (height, cover, biomass, etc.). This approach potentially establishes a powerful baseline and pathway toward developing scalable frameworks and models that can be extended to work with multiple forest attributes at relatively low computational cost compared to more complex deep learning models for Earth observation.

The paper is structured as follows. In Data and Methods, we detail our global training database of using multi-source SAR datasets, modeling framework, and validation strategy using both GEDI and airborne laser scanning (ALS) data. The Results section evaluates model performance across biomes and time periods, while assessing the quality of uncertainty estimates. We then analyze global WSCI fusion predictions and their spatial-temporal patterns, compare results with high-resolution ALS measurements, and demonstrate transfer learning capabilities. We conclude by discussing the benefits, limitations, and potential applications of our approach based on artificial intelligence (AI) for forest management and ecosystem monitoring.

## 2. Data and Methods

**2.1. Multi-source data acquisition and processing framework**

Our methodology integrates three complementary remote sensing data sources to develop and validate a deep learning model for predicting forest structural complexity at high resolution (Figure 1). We set aside 20% of the globally sampled GEDI data for independent validation and used ALS data from two large networks of forest sites to evaluate the model's ability to capture fine-scale complexity patterns.

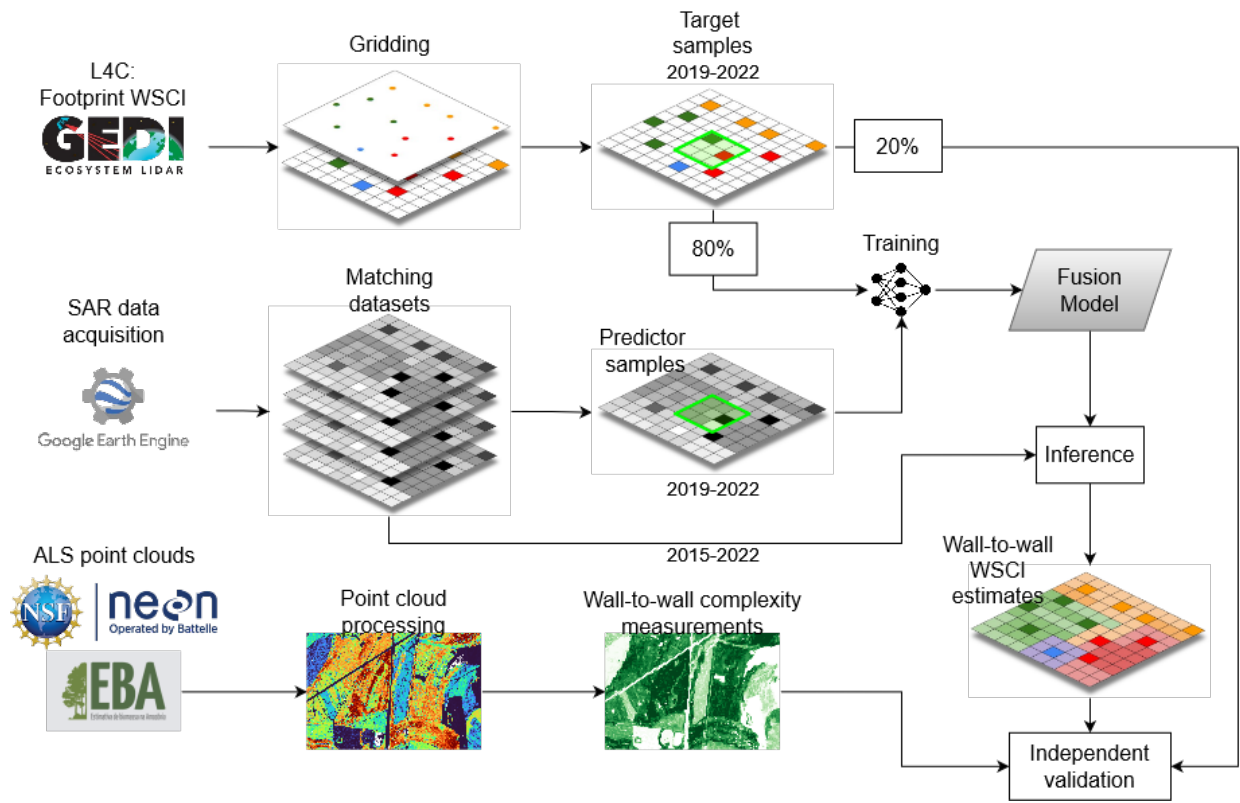

Figure 1. Methodological workflow for the development of a fusion model to map GEDI WSCI at high spatial resolution. The process integrates three primary data sources: GEDI L4C footprint WSCI measurements (top), multi-source SAR datasets (middle), and airborne laser scanning

(ALS) point clouds (bottom). GEDI data are gridded into 1x1 km target samples at 25 m resolution in three-month intervals from April/2019 to December/2022 and separated into training (80%) and validation (20%) sets. SAR data from ALOS-PALSAR and Sentinel-1 sensors are matched spatially with GEDI locations to create predictor variables. The fusion model, trained on these paired datasets, generates wall-to-wall WSCI estimates at 25 m resolution for every quarter between 2015 and 2022. Model performance is assessed through independent validation using both reserved GEDI observations and ALS-derived complexity measurements.

*GEDI Data*

For training the GEDI/SAR WSCI fusion model, we used the GEDI L4C Footprint Level WSCI, Version 2 data product [29]. GEDI footprints were first filtered to retain only high-quality observations and then gridded by averaging WSCI values at a 25 m resolution over 3-month intervals from April 2019 to December 2022. Subsequently, 1×1 km image chips at 25m resolution were randomly sampled across the Earth's land surface. We retained only chips with at least 1% non-empty pixels (i.e., at least 16 out of 1600 pixels) for model training. This 1% threshold balanced data quality with geographic representation, particularly ensuring sufficient inclusion of tropical regions where high-quality GEDI observations are less dense, often below 20 footprints per km² over the entire 2019-2022 time frame [30], while still providing enough samples for the model to learn meaningful spatial patterns.

*SAR Data*

Three, globally available SAR datasets were used (Table 1): the (1) Global PALSAR-2 Yearly Mosaic, Version 2 [31], acquired by the Advanced Land Observing Satellite (ALOS-2) and managed by the Japanese Aerospace Exploration Agency (JAXA), providing L-band measurements at 1270 MHz (23.6 cm wavelength); (2) Sentinel-1 Interferometric Wide Swath (IW) SAR Level-1 Ground Range Detected (GRD) [32], provided by the European Space Agency (ESA), this dataset offers C-band measurements at 5405 MHz (5.6 cm wavelength) with a 12-day temporal resolution; (3) Copernicus DEM [33], a global Digital Surface Model constructed from TanDEM-X data (2010–2015) [34], gap-filled with ancillary products such as SRTM [35] and ASTER [36]. All these datasets were downloaded from Google Earth Engine (GEE) [37].

For the PALSAR mosaics, only pixels classified as land in GEE were used. The analysis ready Sentinel-1 GRD (Ground Range Detected) data available in GEE is pre-processed for noise removal, radiometric calibration, and terrain correction. Mosaics from Sentinel-1 Interferometric Wide Swath (IW) images from both ascending and descending trajectories were generated by averaging backscatter and incidence angle information over 3-month intervals, then resampled to 25 m resolution using bilinear interpolation. The Copernicus DEM was resampled to 25 m with the nearest neighbor method and used as a fixed, non-temporal layer.

Table 1. SAR datasets employed in the current study (GEE = Google Earth Engine, DN = Digital Number, DEM = Digital Elevation Model).

| Dataset | Temporal resolution | Extracted polarizations and other layers | Native resolution on GEE | SAR Band, wavelength |
|---|---|---|---|---|
| PALSAR 1 & 2 | 1 year | HH, HV (16-bit DN, $\gamma^0$) incidence angle (ground) | 25 m | L-band, 23.6 cm |
| Sentinel-1 IW GRD | 3 months | VV, VH (dB, $\sigma^0$) incidence angle (ellipsoid) | 10 m | C-band, 5.6 cm |
| Copernicus DEM | Single observation | DEM (meters above sea level) | 30 m | X-band, 3.1 cm |

We used the PALSAR and Sentinel-1 datasets from 2019 to 2022 as model inputs during training to match GEDI's operational time frame. For inference, the temporal range was extended to 2015–2023 and beyond 52 degrees of latitude in the Northern hemisphere. Signal backscatter information was derived from the single and cross-polarized layers, while incidence angle data were retained to help the deep network learn intrinsic patterns and to flag potential artifacts affecting pixel-level uncertainty. In addition, cyclic encoded geographical coordinates were incorporated as the final three layers in the input stack to aid the model in learning geographical patterns of WSCI distribution.

*ALS Data*

Airborne Laser Scanning (ALS) data was used for independent validation and regional assessments of model performance. The GEDI WSCI product was derived from ALS measurements of 3D canopy entropy ($CE_{XYZ}$) [9] matched to GEDI footprints [13]. This makes the fusion estimates directly comparable to the ALS measurements at 25 m resolution. Wall-to-wall SCI was computed over ALS point clouds from forest sites in the NEON [38] and INPE-EBA [39,40] point cloud networks, which are widely distributed over the United States and Brazilian Amazon regions, respectively. This enabled partial validation of the fusion model and allowed assessment of its adaptation from sparse GEDI shots to continuous ALS measurements. Additionally, 1×1 km image chips from ALS SCI data were sampled to fine-tune the model and evaluate potential performance improvements on a regional scale using both NEON and EBA point clouds.

## 2.2. Neural Network architecture

The selection of neural network architecture and loss function are critical considerations for mapping forest structure efficiently at global scales from sparse observations. An effective architecture must balance multiple competing requirements: it must capture complex spatial relationships in SAR data related to forest structure, maintain computational efficiency for global-scale processing, and accurately quantify prediction uncertainty.

*Model architecture*

We employed an adapted version of the EfficientNetV2 convolutional neural network [41] (Figure 2). EfficientNets deliver comparable or superior performance to other recent CNNs while using significantly fewer trainable parameters [42], thus enabling faster training and inference on large-scale datasets, even in resource-constrained environments.

A key component in the EfficientNet architecture is the Mobile Inverted Bottleneck Convolution (MBConv), which uses inverted depthwise separable convolutions that first expand input channels before capturing spatial information with channel-wise kernels. The EfficientNetV2 introduced Fused Mobile Inverted Bottleneck Convolutions (FusedMBConv) [41], which replace separable convolutions with standard 2D convolutions early in the network, boosting performance at a lower computational cost. Our implementation maintains fixed spatial dimensions throughout the network, preserving the pixel-level context rather than compressing it as in U-Nets [43]. A fixed global normalization layer at the start of the network standardizes all SAR input layers to zero mean and unit variance. Monte Carlo dropout layers enable estimation of model uncertainty via prediction variations over the same inputs [44]. The regression head outputs two channels (mean and variance estimates) using a softplus activation to ensure positive estimates. A 10% negative buffer (4 pixels) is applied to mitigate edge artifacts due to reduced spatial context at image borders.

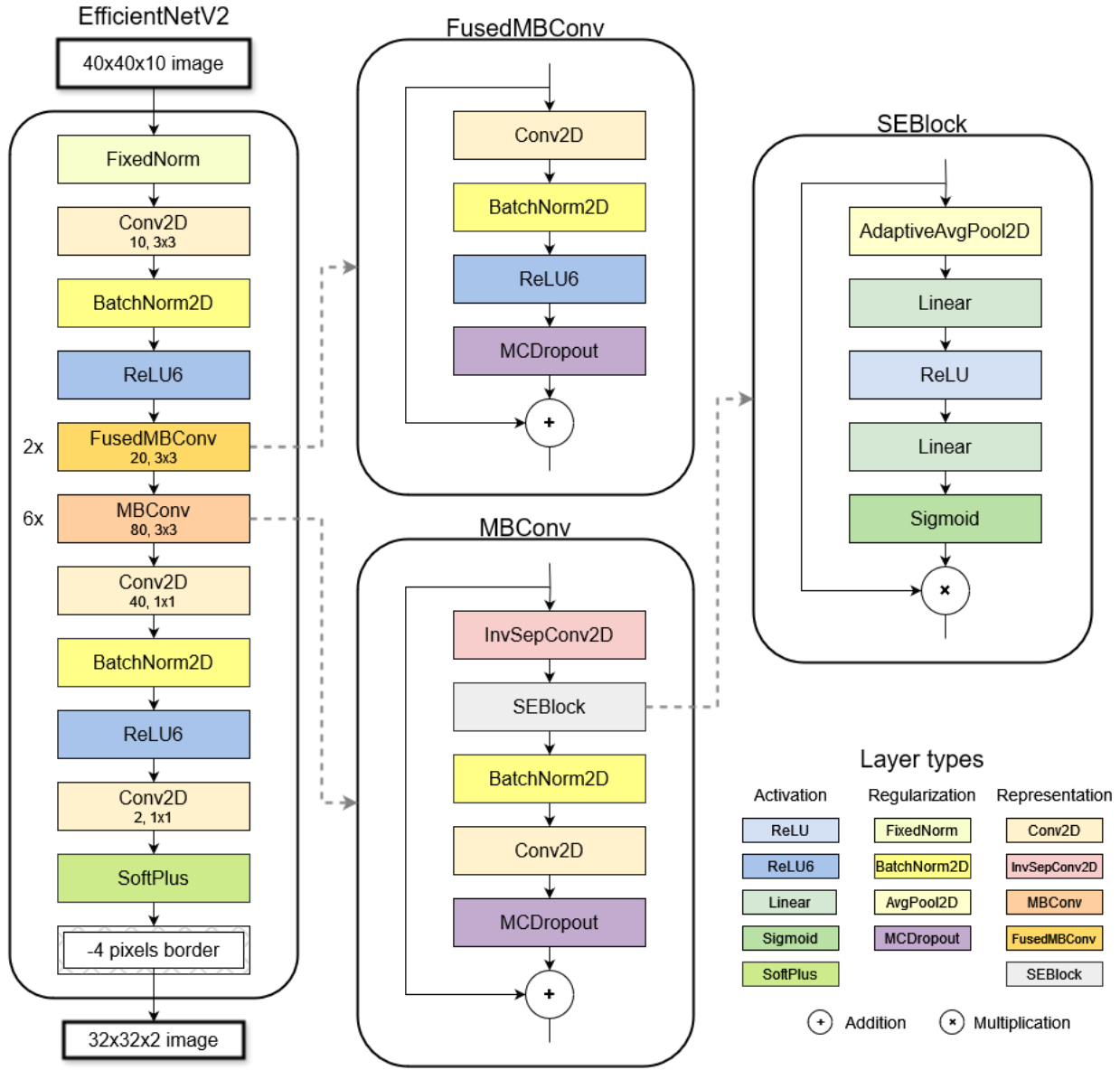

Figure 2. Detailed architecture of our adapted EfficientNetV2 neural network for WSCI estimation. The network (left) processes 40×40×10 input images (7 SAR layers + 3 channels with encoded geographical coordinates) and outputs 32×32×2 predictions (mean and variance) after removing a 4-pixel border to avoid edge artifacts. The network maintains spatial dimensions

throughout processing and features: (1) an initial fixed normalization layer to standardize all channel inputs, (2) two FusedMBConv blocks (top-right), (3) six MBConv blocks with Squeeze and Excitation blocks (bottom-right), and (4) final convolutional layers with softplus activation. Skip connections (represented by the "+" sign) improve gradient flow during training while Monte Carlo Dropout (MCDropout) layers enable model uncertainty estimation. Numbers beside convolution blocks indicate the number of output channels and kernel sizes (e.g., 10, 3×3).

*Loss function*

A masked Gaussian negative log-likelihood (NLL) loss function (Eq. 1) was used in the optimization process, directly enabling the prediction of data uncertainty (i.e. aleatoric variance) alongside WSCI estimates. Because of the negative output buffer, only the central 32×32 pixels of each image chip were used in the NLL computation, while empty pixels with empty targets were excluded from the loss averaging (Eq. 2). This ensures that only pixels intersecting GEDI shots were used to optimize the models' weights.

(1) $NLL = \frac{1}{2}\log(2\pi\sigma^2) + \frac{(y-\mu)^2}{2\sigma^2}$

(2) $Loss = \frac{1}{N_S}\sum_{i \in S}^{n} NLL_i$ where $S = \{i \mid \mu_i > 0\}$

Where: y = model prediction; µ = target value (GEDI WSCI); $\sigma^2$ = variance; S = set of valid pixels (recorded GEDI WSCI); $N_S$ = number of elements in S.

*Model training*

The image chips comprising the training dataset were partitioned into spatial blocks approximately 80×80 km in size. Each block contained up to 300 samples of 1×1 km image chips at 25 m resolution, randomly sampled in 3-month intervals between April 2019 and December 2022. Each sample consists of an input stack with dimensions 10×40×40 (7 SAR layers + 3 geographical coordinate layers) and a corresponding 1×40×40 target of rasterized GEDI WSCI values averaged over the respective 3-month period. Only samples with valid data across all layers were included. We split the samples into two independent sets, where 80% of the spatial blocks were used for training while 20% were reserved for testing. The model was optimized using the Adam optimizer [45], with a multi-step learning rate scheduler reducing the learning rate by a factor of 0.1 at 10%, 20%, and 50% of the training epochs.

*Ensemble predictions and uncertainty estimation*

During inference, 40×40 km tiles were used to generate prediction batches of 1600 samples. Each tile underwent model prediction five times with a slight offset in both longitude and latitude directions, ensuring each pixel was predicted from multiple overlapping windows with different spatial contexts. Ensemble predictions were obtained by averaging these five estimates. This approach reduced edge artifacts between adjacent chips on inference, preserving the spatial continuity between them, and combined with Monte Carlo dropout, allowed us to derive epistemic (model) uncertainty as the standard deviation of mean WSCI across passes for every pixel. The joint uncertainty was then computed as the total standard deviation from the combined aleatoric and epistemic variances (Eq. 3). The validation of uncertainty estimates is addressed in the following section.

$$(3) \quad \sigma_{total} = \sqrt{\sigma_{data}^2 + \sigma_{model}^2}$$

Where: $\sigma_{total}$ = total standard deviation; $\sigma_{data}^2$ = aleatoric variance; $\sigma_{model}^2$ = epistemic variance.

## 2.3. Model validation

*GEDI independent validation*

Ensemble predictions were extracted from all image chips in the test set and compared, pixel-by-pixel, with corresponding GEDI WSCI observations. Performance was evaluated both geographically, by comparing predictions with GEDI observations across different biomes [46], and temporally, by assessing consistency over various time periods. Root mean squared error (RMSE), bias, and $R^2$ metrics were used to quantify model accuracy and precision.

*ALS independent validation*

The GEDI L4C product provides uncertainty estimates as prediction intervals, however we did not use these estimates to train the fusion model and considered the GEDI WSCI estimates as true targets when training our EfficientNetV2. To compensate for this assumption, we used samples of ALS wall-to-wall measurements of structural complexity, which are directly calculated and do not have associated model uncertainty and low measurement error, thus providing spatially continuous samples that enable further validation against ground truth information.

The NEON and EBA datasets were thus considered as ground truth source of structural complexity for regional validation. While GEDI provides discrete footprint measurements, the airborne lidar datasets provide wall-to-wall coverage that allows for more comprehensive spatial analysis. We compared ALS-derived 3D canopy entropy ($CE_{XYZ}$) measurements at 25 m resolution, which is the source index from which WSCI was created, with corresponding model predictions to assess the fusion model's ability to (1) learn continuous structural complexity patterns from sparse GEDI observations, (2) accurately estimate WSCI for time periods outside the training dataset (2015-2018) and, (3) capture local gradients and fine-scale patterns of structural complexity. Beyond standard performance metrics (RMSE, bias, R²), we evaluated the model's ability to preserve spatial correlation patterns by calculating Moran's I spatial cross-correlation and residual correlation between predicted and observed structural complexity at each ALS site. This analysis is particularly important for assessing whether the model successfully filled gaps between GEDI footprints while maintaining realistic spatial structure.

*Uncertainty validation*

The fusion model predicts both the WSCI value and its uncertainty at each pixel. To validate these uncertainty estimates, we assessed their calibration using coverage analysis. In a well-calibrated model with Gaussian error distribution (as assumed by our NLL loss function), approximately 68% of true values should fall within one standard deviation of the predicted mean, and 95% within two standard deviations. For each prediction-observation pair, we calculated Z-scores (Eq. 4).

(4) $Z = \frac{y - \mu}{\sigma_{total}}$ where $Z \sim N(0,1)$

Where y is the predicted mean, μ is the observed GEDI WSCI value, and $\sigma_{total}$ is the predicted standard deviation. If uncertainty estimates are well-calibrated, these Z-scores should follow a standard normal distribution.

We conducted this coverage analysis across different WSCI value ranges, biomes, and time periods using both GEDI and ALS validation data. This approach allowed us to identify potential patterns of overconfidence (where predicted uncertainty is too small, i.e., overly optimistic) and underconfidence (where predicted uncertainty is too large, i.e., overly conservative) in the model's estimates.

## 2.4. Feature importance

*Recombination of model inputs*

To determine the contribution of individual data sources, we trained several versions of the EfficientNetV2 model on a reduced sample (~800,000 GEDI shots, with 20% set aside for validation), each excluding different input layers. The performance of model trials was then compared using RMSE, bias and $R^2$ metrics.

*Deep SHAP explainer with global background*

SHAP (SHapley Additive exPlanations) analysis [47] was performed on the fully trained model to evaluate the relative contribution of each input layer. A Deep SHAP explainer was applied to randomly sampled pixels worldwide, calibrating each prediction against a global background of 100 image chips. SHAP values were then summarized spatially and channel-wise (Figure 3), quantifying the influence of the spatial neighborhood as a function of the distance between predictor and predicted pixels, and to assess the relative importance of different input channels (SAR bands, geographical coordinates). Additionally, we checked for temporal and geographical patterns in the SHAP explanations through latitudinal temporal trends and biome-wide patterns of accumulated SHAP importance from different data sources.

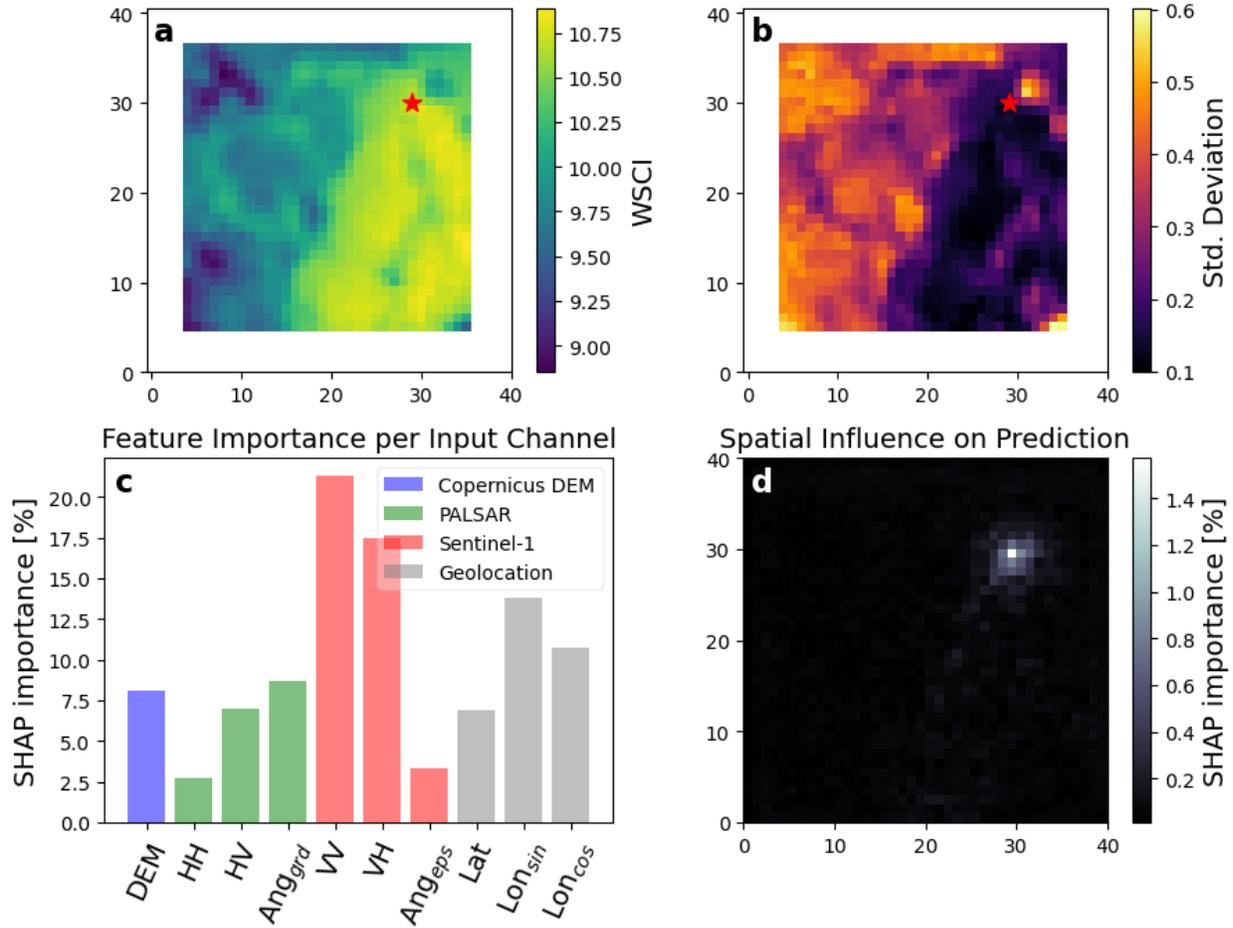

Figure 3. Spatial and channel-wise feature importance analysis using Deep SHAP explainers for a single predicted pixel. This visualization illustrates the various factors influencing the model's prediction at the individual pixel level. (a) Wall-to-wall WSCI fusion values for a 1×1 km image chip at 25m resolution, with the red star marking the target pixel being analyzed. (b) Corresponding standard deviation map showing prediction uncertainty across the same area. (c) Relative contribution of each input channel to the prediction at the target pixel. (d) Spatial influence map showing how each pixel in the 40×40 pixel window (1 km$^2$) affects the target prediction, revealing a strong distance-decay relationship where nearby pixels exert substantially greater influence than distant ones, highlighting the ability of CNNs to capture local neighborhood context while minimizing influence from distant pixels, thus handling forest edge effects implicitly.

2.5. Fine tuning and transfer learning

To evaluate regional adaptability, the model was fine-tuned using ALS $CE_{XYZ}$ measurements as targets instead of GEDI data. Two strategies were employed: (1) re-optimizing all pre-trained model parameters and (2) retraining only the last convolutional layer in the network (the model head, Figure 2), freezing all the other weights. Improvements were assessed using RMSE, bias, $R^2$, and uncertainty coverage analyses.

Since the GEDI WSCI integrates various structural attributes and is highly correlated with other GEDI measurements (e.g., canopy height, canopy cover, foliage height diversity, aboveground biomass) [13], we expect that a model trained on WSCI can be adapted to predict these related variables. We re-trained the model to predict GEDI RH98 (canopy height) and canopy cover using globally sampled targets. Both transfer learning experiments were carried out using full weight updates and frozen-weights strategies.

## 3. Results

The EfficientNetV2 model was trained on approximately 5 million 1×1 km image chips randomly sampled across the GEDI geographical domain (52° North to 52° South), representing ~133 million GEDI footprints. Our final model architecture contains 365,682 trainable parameters and required approximately 200 hours of training on an NVIDIA Quadro P4000 8GB GPU (50 epochs, batch size = 96, initial learning rate = 0.001, dropout rate = 20%, processing rate ~2 batches per second).

### 3.1. Model performance

The trained model demonstrated strong performance when evaluated against independent test data, capturing 82% of the variability in GEDI WSCI observations with minimal bias (Figure 4a). Beyond accuracy metrics, we evaluated the quality of uncertainty estimates through coverage analysis. Our model achieved a coverage of 71% for one standard deviation (Figure 4c), indicating slightly conservative but well-calibrated uncertainty estimates. This calibration remained relatively stable across most WSCI values (65%–80%), though the model tended to be somewhat overconfident (coverage below 69%) around WSCI = 9 and more conservative at extreme values (Figure 4d). The model showed moderately positive correlation between predicted uncertainty and actual residuals (Figure 4b), further confirming appropriate uncertainty estimation. We note that the model could not estimate WSCI values below 7.4, which typically represent unforested or sparsely vegetated areas outside the model's training domain.

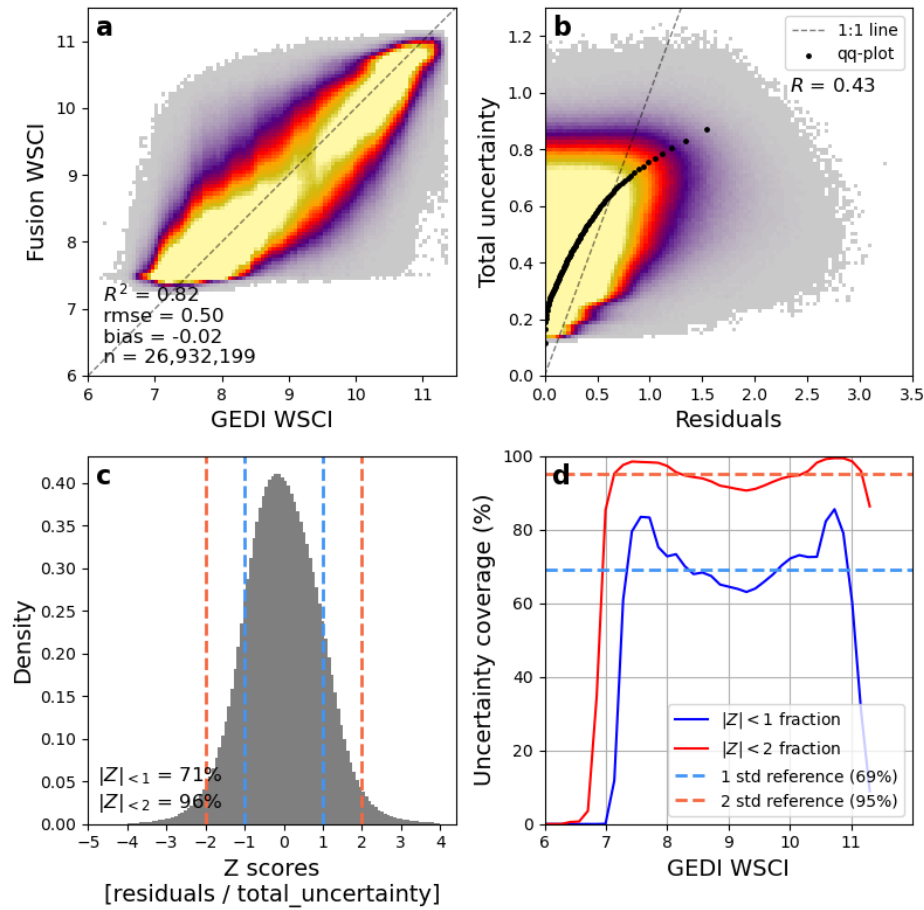

Figure 4. Performance evaluation of the GEDI-SAR WSCI fusion model using independent test data. (a) Density scatterplot showing the relationship between predicted fusion WSCI and observed GEDI WSCI values ($R^2 = 0.82$, RMSE = 0.50, bias = -0.02, n = 26,932,199). (b) Relationship between model uncertainty estimates and absolute residuals, showing positive correlation ($r = 0.43$), and underestimation of model uncertainty above 0.7 standard deviations highlighted by the qq-plot. (c) Distribution of normalized residuals (Z-scores), calculated as residuals divided by predicted total uncertainty, with vertical lines indicating standard deviation coverage (71% and 96% for 1 and 2 standard deviations, respectively). (d) Uncertainty coverage analysis across different WSCI intervals, showing the percentage of observations falling within one standard deviation (blue line) and two standard deviations (red line) of predictions. Dashed reference lines indicate the expected coverage for well-calibrated uncertainties (69% and 95% respectively).

## 3.2. Temporal and Geographical Consistency

The fusion model exhibited temporal stability throughout the GEDI operational period (Sup. Fig. 1), with performance metrics fluctuating by less than 5% across quarterly intervals from 2019 to 2022. Geographically, performance varied systematically across biomes (Sup. Table 1, Figure 5), with tropical moist broadleaf forests showing the highest explanatory power ($R^2 = 0.75 \pm 0.03$) and lowest error (RMSE = $0.47 \pm 0.03$). Most forested biomes maintained $r^2$ values exceeding 0.64 (Sup. Table 1), while uncertainty calibration remained consistently near the theoretical optimum (|Z-score| < 1 fraction ≈ 0.70). Performance declined in regions with sparse vegetation (deserts/xeric shrublands, $R^2 = 0.48 \pm 0.05$) and areas with limited GEDI coverage (boreal forests/taiga, $R^2 = 0.59 \pm 0.06$). Summarizing performance by season (Sup. Fig. 2) revealed optimal model performance from April to September, coinciding with leaf-on conditions in Northern Hemisphere temperate forests and reduced cloud cover during tropical dry seasons.

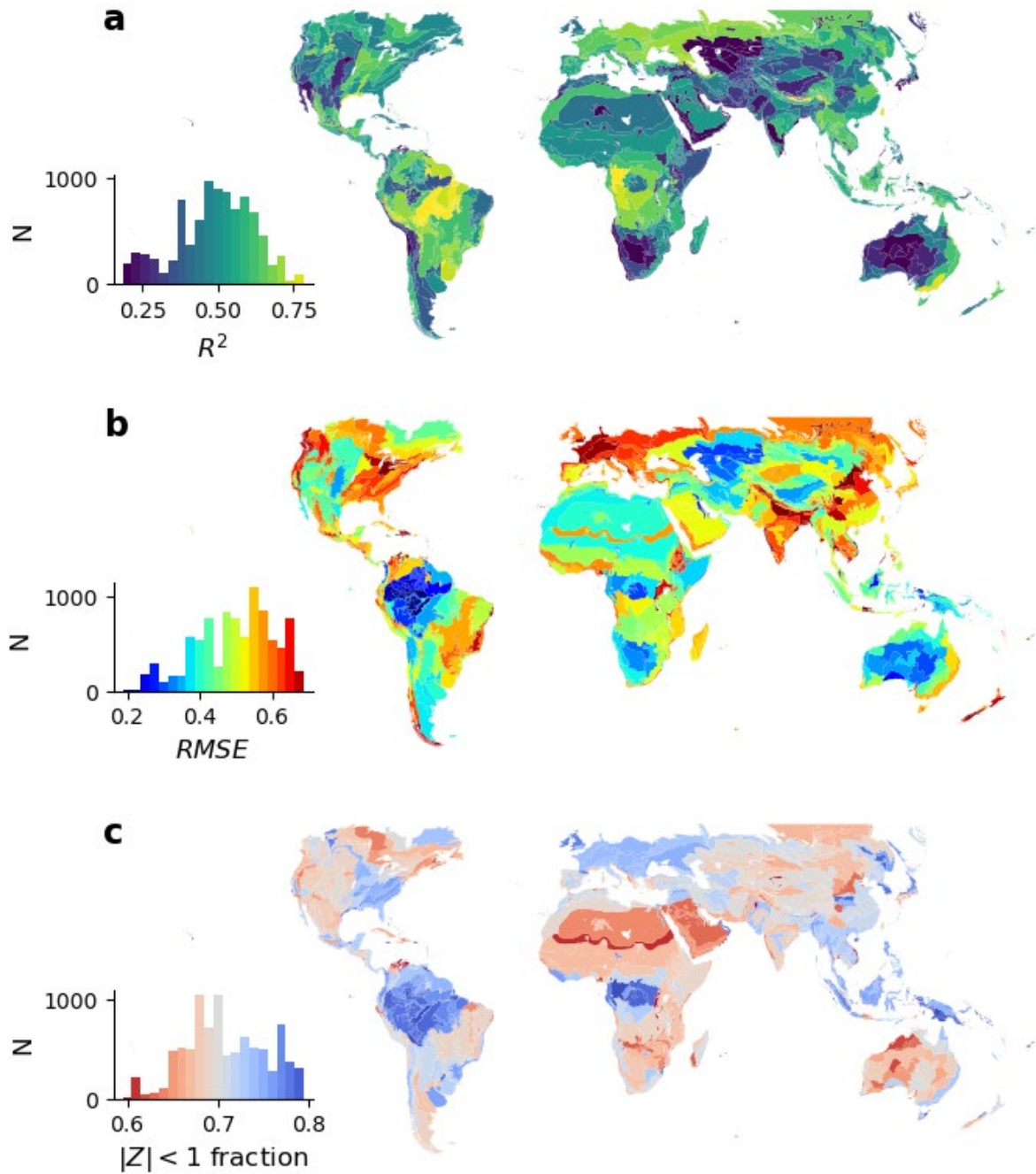

Figure 5. Spatial distribution of model performance metrics across ecoregions [46] within the GEDI coverage area (51.6°N to 51.6°S). Maps show (a) coefficient of determination ($R^2$), (b) root mean square error (RMSE), and (c) uncertainty coverage (fraction of observations falling within one standard deviation of predictions) by ecoregion for validation data from April 2019 to December 2022. Histograms show the frequency distribution of each metric across all ecoregions.

## 3.3. Global mapping and temporal trends

Following model training and validation, we deployed the fusion model for global inference to generate wall-to-wall WSCI maps (Figure 6). The inference process was conducted on overlapping 40×40 km image tiles (batch size = 1600), balancing CPU efficiency and memory usage, with each tile requiring ~300 seconds on a single Intel Xeon Gold 6148 CPU core (2.40 GHz). This processing approach was used to predict WSCI wall-to-wall globally, extending mapping beyond GEDI's geographical domain, producing complete global land coverage including previously unmapped regions such as boreal forests (Figure 6). Each global quarterly map consists of approximately 120,000 image tiles, requiring ~10,000 CPU hours of processing time (equivalent to 1 day using 420 CPUs). We generated maps for all quarters between January 2015 and December 2022, creating a consistent time series that includes periods before GEDI's operational timeline.

The uncertainty map (Figure 6, bottom) reveals patterns in prediction confidence across landscapes. Uncertainty estimates were typically lowest in areas with extreme WSCI values (either sparse vegetation or densely forested regions) where structural conditions are more homogeneous. In contrast, higher uncertainty was observed in transitional zones, forest edges, and areas where the input SAR mosaics showed spatial discontinuities or poor harmonization. The high-resolution insets in Figure 6 illustrate the model's ability to capture fine-scale spatial patterns in various regions. All fusion training samples included data from both PALSAR and Sentinel-1 mosaics, therefore model predictions were not generated for areas where one of these datasets was missing entirely at any given time (e.g. empty stripes in Northern Russia in Figure 6).

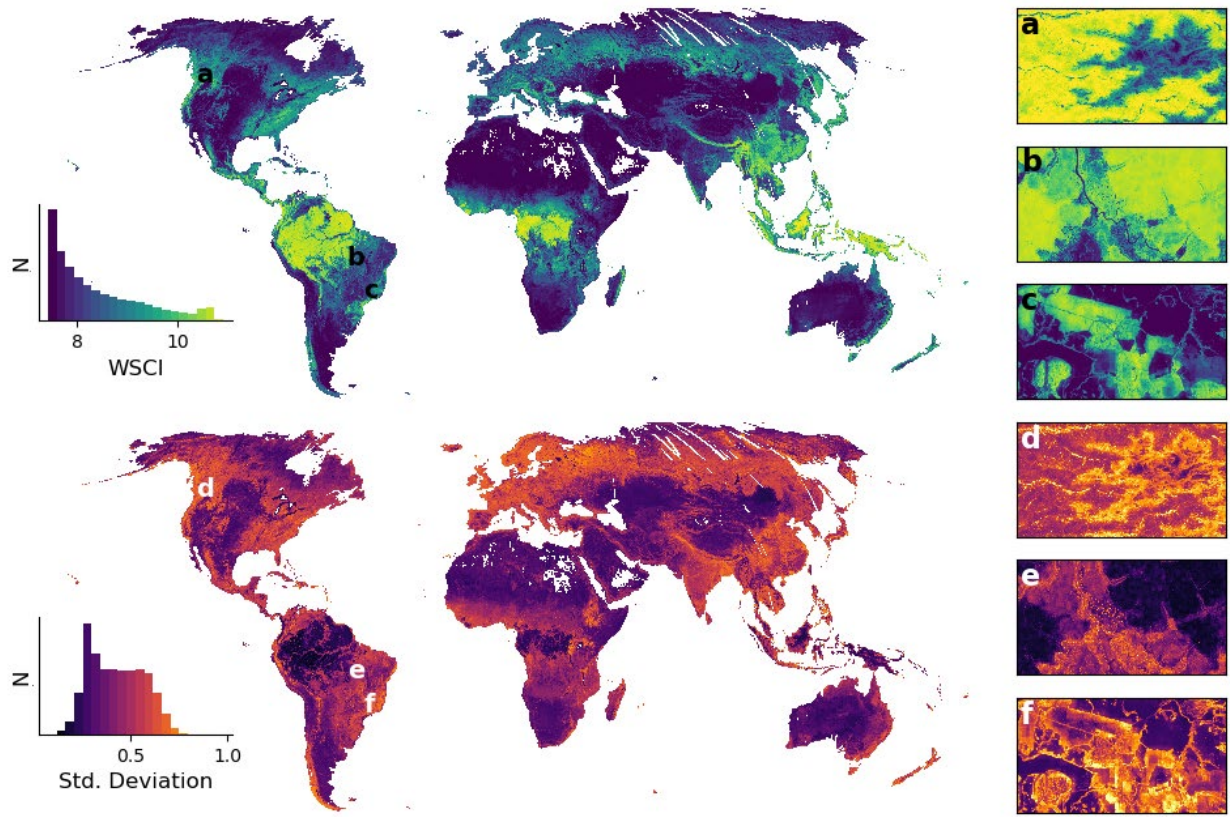

Figure 6. Global map of Waveform Structural Complexity Index (WSCI) and associated uncertainty produced by the GEDI-SAR fusion model for April-June 2019-2022. The top panel shows predicted WSCI values while the bottom panel displays the predicted standard deviation. These global inference maps extend beyond GEDI's orbital coverage and include the Earth's entire landmass. High-resolution insets highlight: (a,d) structural complexity along a topographical gradient in Olympic National Park, USA; (b,e) fire disturbed forest in the Xingu Indigenous Territory in the southern Amazon; and (c,f) a mosaic of planted and natural forest patches in Brazil's Atlantic rainforest. Note how transitional areas show elevated predicted uncertainty. White stripes in Northern Russia are sections where either PALSAR or Sentinel-1 data were entirely missing when generating the image mosaic.

Our study period () revealed seasonal patterns in forest structural complexity, particularly at higher latitudes (Figure 7, top panel), with higher variability of model predictions within the same year and less structured temporal patterns prior to 2017. While no significant long-term trends were detected, we observed seasonal cycles in temperate and boreal forest regions (most notably above 23° latitude in the Northern hemisphere), with complexity increasing during spring and summer months. This pattern aligns with the seasonal leaf development in deciduous forests during growing seasons, as well as snow and freeze/thaw cycles in the boreal zone that may affect the SAR signal.

Data uncertainty (aleatoric standard deviation, Figure 7, middle panel) showed strong latitudinal patterns. Uncertainty was consistently lowest near the equator, where evergreen tropical forests maintain relatively stable, complex structures year-round. In contrast, higher and more variable data uncertainty was observed at temperate and boreal latitudes (beyond 23° latitude), reflecting the greater structural variability and seasonal dynamics in these regions. Model uncertainty (epistemic standard deviation, Figure 7, bottom panel) revealed two key patterns. First, we observed higher uncertainty values before the end of 2016, corresponding to over 2 years before the GEDI period and the pre-Sentinel-1B period when only single-satellite data was available. The dual-satellite configuration during model training (2019-2022) likely provided more consistent input features, reducing prediction uncertainty. However, Sentinel-1B ceased operations in December 2021, reducing our model to single-satellite data for Sentinel-1 with limited concurrent GEDI observations for the final training year. This temporal non-stationarity in satellite configurations may contribute to elevated uncertainty estimates when extrapolating several years beyond the training period. Additionally, elevated uncertainty in 2015-2016 may reflect early operational issues in PALSAR-2 data [48,49] that may have persisted despite reprocessing with improved calibration algorithms implemented in late 2016 [50], as well as extrinsic factors such as surface moisture conditions. Second, model uncertainty increased considerably beyond GEDI's coverage limits (51.6° latitude), where no training data was available, reflecting less confident extrapolations in these regions. Lower uncertainty estimates were associated to extreme values in the predicted WSCI range, with lowest average uncertainty observed near the Equator, where high complexity forests are concentrated, and above 70° latitude, where lowest average WSCI occurs (Figure 7).

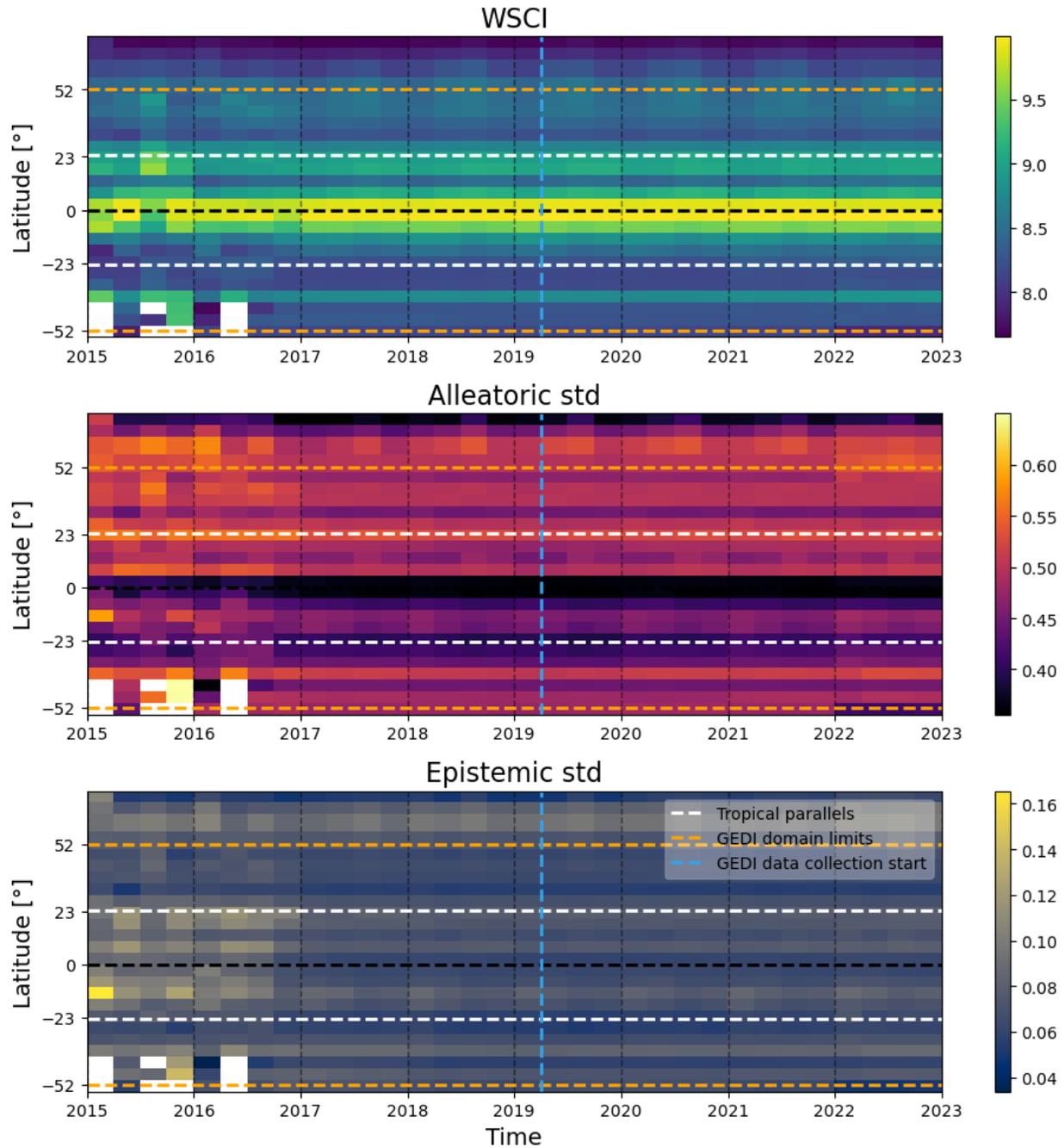

Figure 7. Latitudinal patterns in WSCI and estimated uncertainty components from 2015 to 2022. Data are averaged in 5-degree latitude bands and 3-month intervals. Top panel: Predicted WSCI values showing seasonal cycles at temperate and boreal latitudes. Middle panel: Data uncertainty (aleatoric standard deviation), with lower values in tropical regions. Bottom panel: model uncertainty (epistemic standard deviation), showing higher values before 2017 (pre-Sentinel-1B deployment) and outside GEDI's coverage limits.

## 3.4. Feature Importance and Sensitivity Analysis

We hypothesize that multi-frequency SAR data combined with geographical context inputs improves structural complexity predictions by capturing complementary aspects of forest structure, with different wavelengths interacting with distinct canopy layers while geographical and topographical information provide further spatial context. To test this hypothesis, we analyzed model performance with varied input combinations (Figure 8) and conducted feature importance assessments through SHAP explainers over predictions from the model used for global inference (Figure 9). The full model incorporating all inputs achieved highest accuracy ($R^2 = 0.81$) and lowest error (RMSE = 0.55) while producing unbiased estimates (bias = -0.01), with models using only L-band PALSAR data ($R^2 = 0.69$) substantially outperforming C-band Sentinel-1 models ($R^2 = 0.46$). This performance likely difference reflects the physical properties of SAR wavelengths: L-band signals penetrate deeper into forest canopies to interact with branches and stems, while C-band signals primarily reflect from upper canopy foliage [18]. Removing either topographical or geographical context reduced performance, highlighting their importance for enforcing consistency in model predictions across different geographical regions globally, i.e. spatial regularization (Figure 8).

Despite their weaker performance in isolation, SHAP analysis revealed Sentinel-1 features contributed most significantly to final model predictions (Figure 9b), likely due to their higher temporal resolution in our training database (Table 1), providing more exact temporal matching with GEDI observations and enabling inference in 3-month intervals. The geographical distribution of feature importance showed regional specialization (Figure 9a), with PALSAR more influential in tropical and temperate forests and DEM data contributing substantially in mountainous regions where topography constrains forest development [51,52]. A strong contrast between global and local feature importance was also observed (Figure 9b,c). Geographical coordinates showed high importance globally but minimal influence within the same region, confirming their role as a spatial prior that establishes region-specific baselines, an important function for geographical generalization in large-scale deep learning models (Reichstein et al. 2019; Tang et al. 2015; Lang et al. 2023). These results support our hypothesis that combining data sources sensitive to different canopy components, enhanced with geographical and topographical context, likely enables more comprehensive characterization of three-dimensional forest structure.

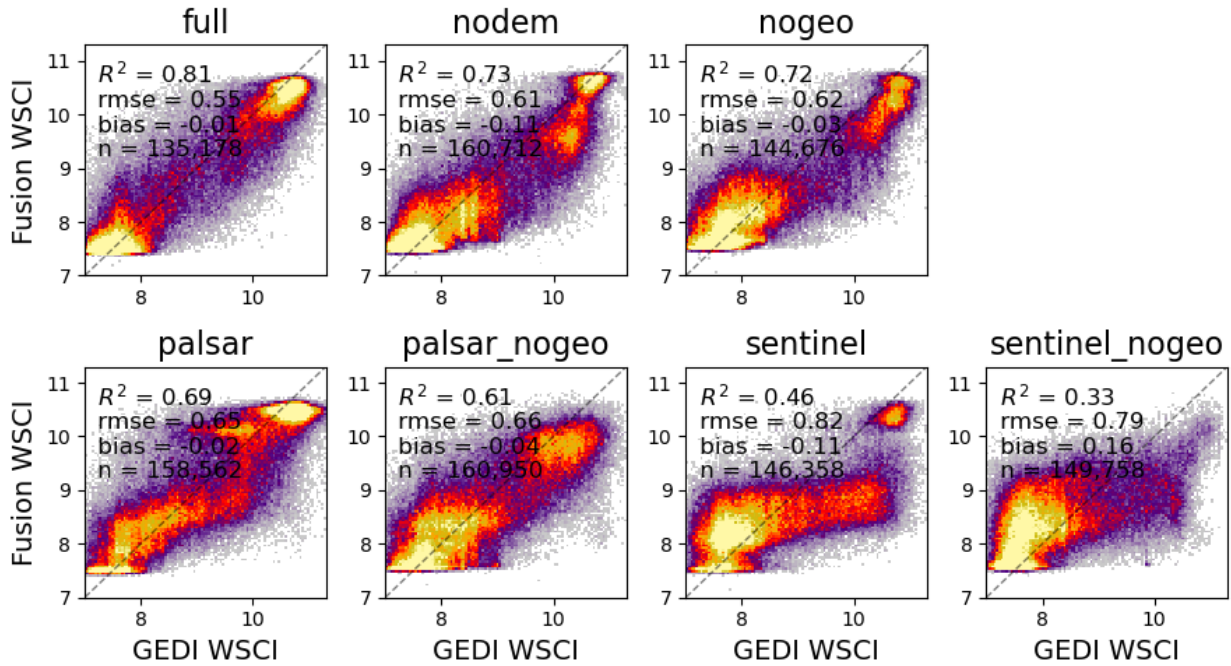

Figure 8. Model performance comparison using different combinations of input data to predict GEDI WSCI. Density scatterplots show the relationship between predicted and observed values for models with: all inputs (full), without topography (nodem), without geographical coordinates (nogeo), only L-band PALSAR, PALSAR without coordinates (palsar_nogeo), only C-band Sentinel-1, and Sentinel-1 without coordinates (sentinel_nogeo). Performance metrics ($R^2$, RMSE, bias, sample size) are shown in each panel.

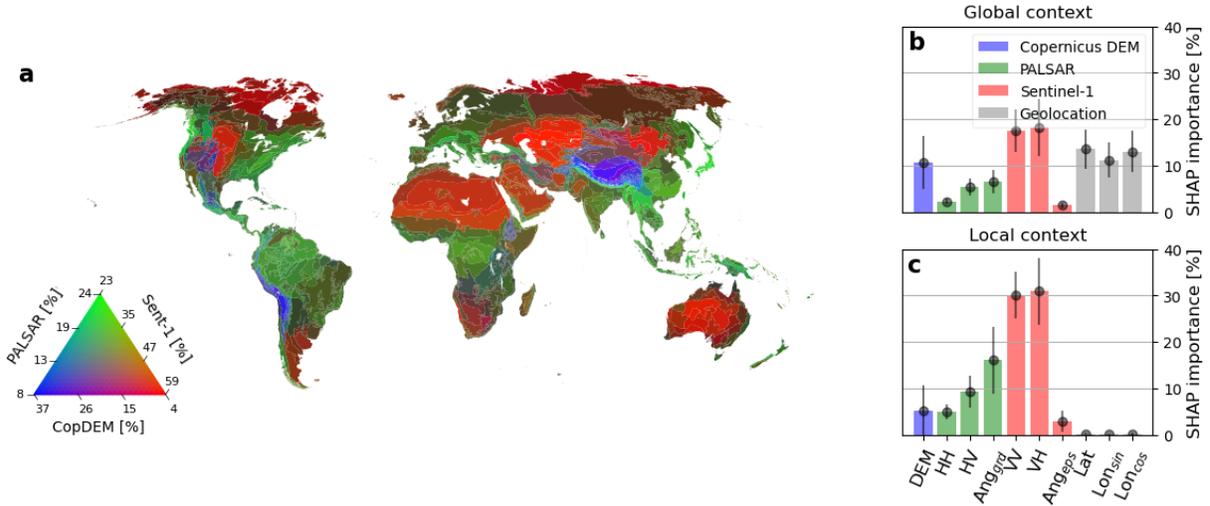

Figure 9. SHAP-based feature importance analysis for the GEDI-SAR WSCI fusion model. (a) Map showing the relative contribution of different SAR datasets to model predictions by ecoregion, with RGB color scheme representing Copernicus DEM (blue), PALSAR (green), and Sentinel-1 (red) contributions. (b) Global feature importance showing mean and standard deviation of SHAP values for each input layer across geographically random samples. (c) Local feature importance showing SHAP values when analyzing predictions within the same geographical region, highlighting the reduced importance of geographical coordinates at local scales. Longitude values were cyclic encoded using sine and cosine functions. Bright green areas represent near-maximum contribution of PALSAR bands, while bright red areas represent near-maximum contribution of Sentinel-1.

No temporal trends were observed in feature importance except for seasonal variations at higher latitudes (Sup. Fig. 3a). In these areas, the model relied less on Sentinel-1 predictors during spring and summer months, while the importance of geographical coordinates increased, aligning with seasonal patterns seen in aleatoric uncertainty (Figure 7). This suggests that the model compensates for increased uncertainty in Sentinel-1 backscatter by relying more heavily on geographical context information during these periods. Separately, analysis of spatial prediction patterns within 1×1 km windows (Figure 3d) revealed that model predictions are most strongly influenced by immediate neighboring pixels, with influence decreasing with distance and becoming negligible beyond approximately 300 meters (Sup. Fig. 3).

### 3.5. Validation with ALS data

When compared to ALS structural complexity measurements (CE$_{XYZ}$), the global model explained 47% its variability over NEON sites (Figure 10a) and benefited markedly from fine-tuning (Sup. Fig. 4), which modestly improved R² (from 0.47 to 0.56), RMSE (from 0.97 to 0.90), bias (from -0.28 to -0.11 and further expanded the prediction range of the model down to WSCI = 6.5. At the EBA sites (Figure 10b), fine tuning did not improve model performance (Sup. Fig. 5). These contrasting responses demonstrate that while our global model performs adequately across biomes, fine-tuning with ALS may help to further improve model performance in some regions.

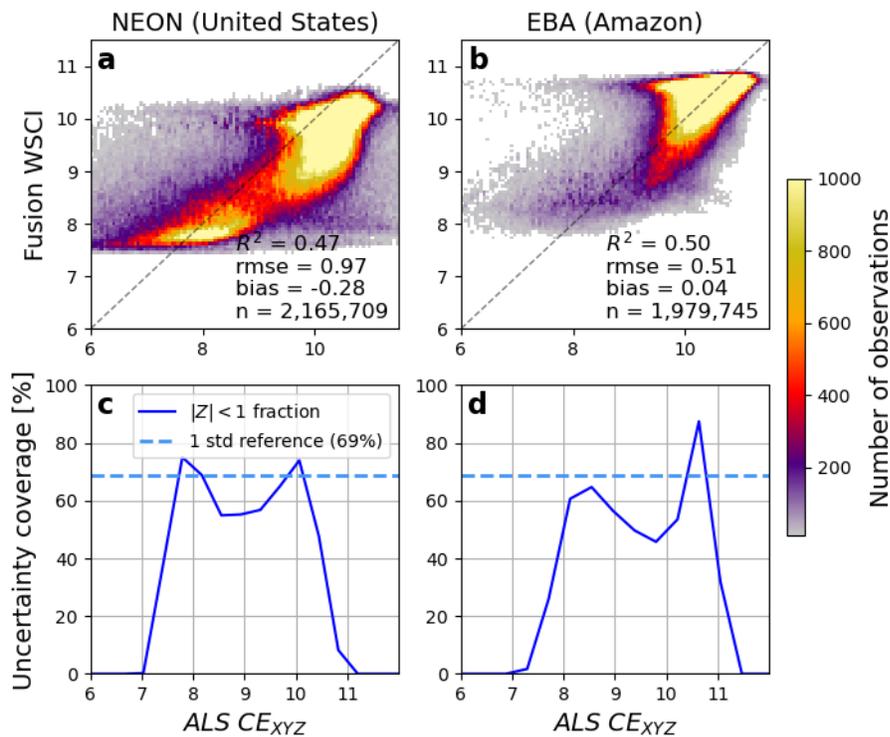

Figure 10. Model performance of the GEDI-SAR WSCI Fusion against structural complexity measurements (CE$_{XYZ}$) from ALS data in the (a,c) NEON network over the United States and (b,d) EBA network in the Brazilian Amazon rainforest.

Spatial correlation analysis reveals that our fusion model's effectiveness varies systematically with site-level patterns of forest structural complexity (Figure 11). In structurally heterogeneous sites, particularly NEON sites across the US, the model achieves strong spatial cross-correlation (I > 0.4) between predicted WSCI and ALS-derived CE$_{XYZ}$ measurements (Figure 11a). This demonstrates that the model successfully captures broad spatial patterns when distinct structural gradients provide clear learning signals. Conversely, in more homogeneous forest environments, as in the Amazon EBA sites, spatial cross-correlations are weaker (I < 0.4), indicating reduced model performance where structural variation is subtle.

However, analysis of spatial autocorrelation in model residuals reveals persistent limitations in fine-scale accuracy (Figure 11b). The presence of spatially structured prediction errors, particularly pronounced in homogeneous Amazon sites, indicates that the model fails to capture local spatial patterns within structurally uniform areas. Visual inspection of representative sites confirms this pattern (Figure 12): in heterogeneous forests, the model successfully reproduces major spatial transitions between distinct forest patches, but in homogeneous areas, it captures overall complexity magnitude while missing finer-scale spatial variation. Prediction uncertainties also show spatial organization, with elevated values along forest edges and transition zones, particularly in the global model (Sup. Fig. 6).

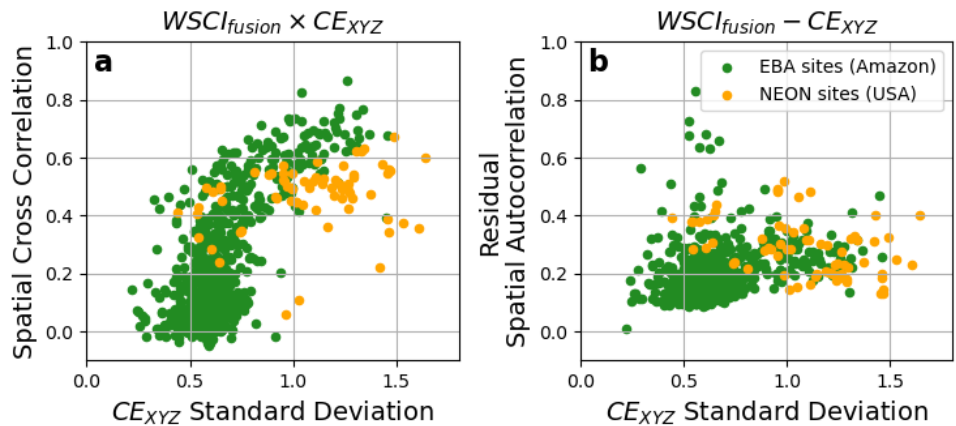

Figure 11. Spatial correlation analysis comparing fusion model predictions with ALS measurements across individual forest sites. (a) Cross-correlation between predicted WSCI and observed $CE_{XYZ}$ plotted against the structural heterogeneity ($CE_{XYZ}$ standard deviation) of each site. Higher correlation in more heterogeneous sites indicates the model successfully captures meaningful structural gradients when present. (b) Spatial autocorrelation of model residuals versus site $CE_{XYZ}$ heterogeneity, showing some remaining spatial structure in errors, particularly in less heterogeneous sites in the Amazon. Green points represent EBA sites in the Amazon, while yellow points represent NEON sites across the continental United States.

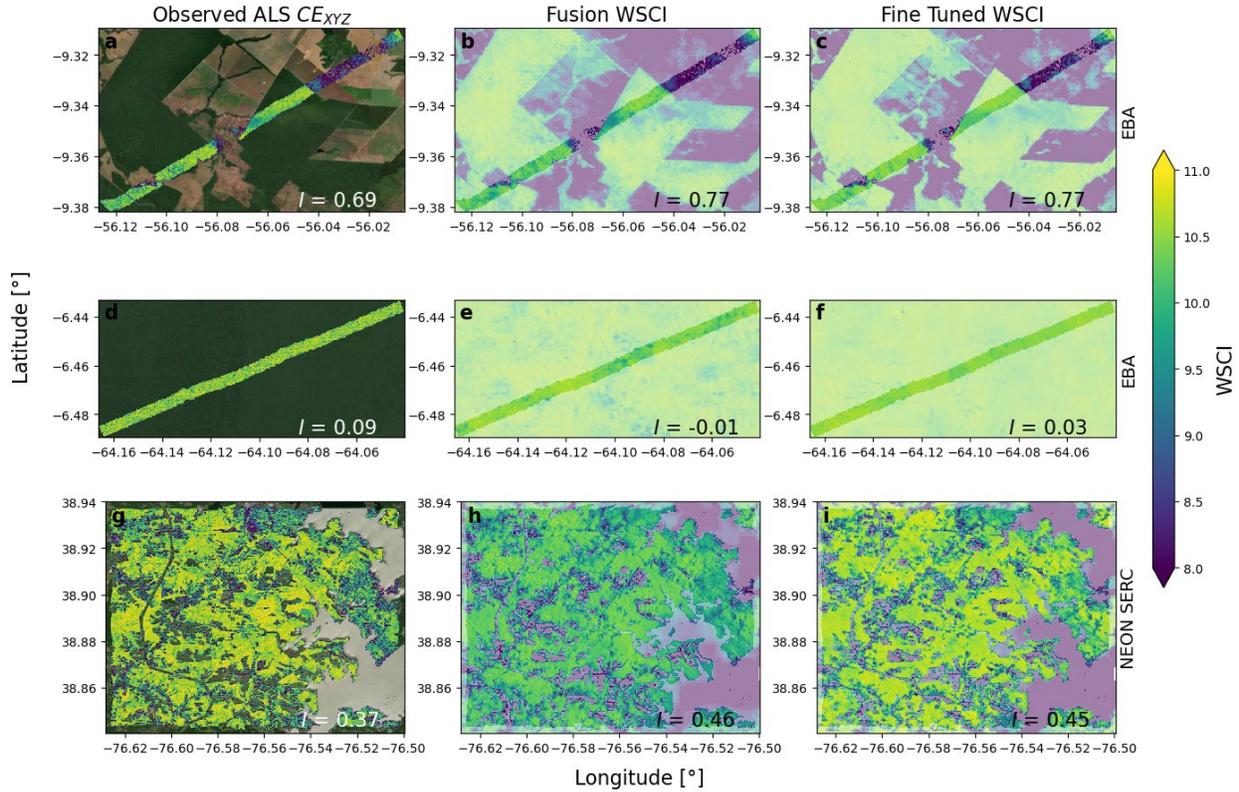

Figure 12. Wall-to-wall structural complexity maps at 25 m resolution for individual sites. Rows show: two EBA sites in the Brazilian Amazon (top, middle) and the NEON SERC site on 2019 (bottom). Brighter areas in the images represent areas covered by ALS. Columns display: (a,d,g) ALS-derived $CE_{XYZ}$ measurements, (b,e,h) predictions from the global fusion model, and (c,f,i) predictions after region specific fine-tuning of the model. Spatial correlations (Moran's I) in the bottom right of each panel represent autocorrelation of $CE_{XYZ}$ observations on the left column (a,d,g), and cross-correlation between predicted and observed structural complexity in the center and right columns, demonstrating how well each model captures the spatial patterns present in the reference data.

### 3.6. Transfer learning

Leveraging the integrative nature of GEDI WSCI, we extended our model to predict two additional forest metrics: canopy height (RH98) and canopy cover, using approximately 50 million GEDI footprints (Figure 13). We evaluated two transfer learning strategies: fully updating all model weights versus freezing the feature extraction layers and only updating the regression head of our neural network (Figure 2). Both approaches demonstrated moderate performance, capturing 69% of canopy height variability (Figure 13a and Sup. Fig. 7) and 66% of canopy cover variability (Figure 13b and Sup. Fig.8). The fully transferred model converged after 20 epochs for canopy height and 37 epochs for canopy cover, while the frozen-weights approach required 27 epochs for height and only 8 epochs for cover. Despite similar performance metrics, the frozen-weights model produced slightly less biased estimates for both variables (Sup. Fig. 7 and Sup. Fig. 8) and offered higher computational efficiency, processing batches four times faster by computing gradients only for the model's regression head. This efficiency advantage makes the frozen-weights approach particularly attractive for operational applications or when computational resources are limited. Predictions saturated around 36 m for canopy height, suggesting the input data may lack sensitivity to identify very tall forests or that additional adjustments to hyperparameters or custom loss functions with specific regularization criteria would be needed to accurately capture extreme canopy height values [14], or that uniform sampling of different intervals of the target variable should be considered for training the model [15]. Nevertheless, these results demonstrate that a single foundational model architecture can be used as a flexible approach to predict multiple forest structural attributes through transfer learning.

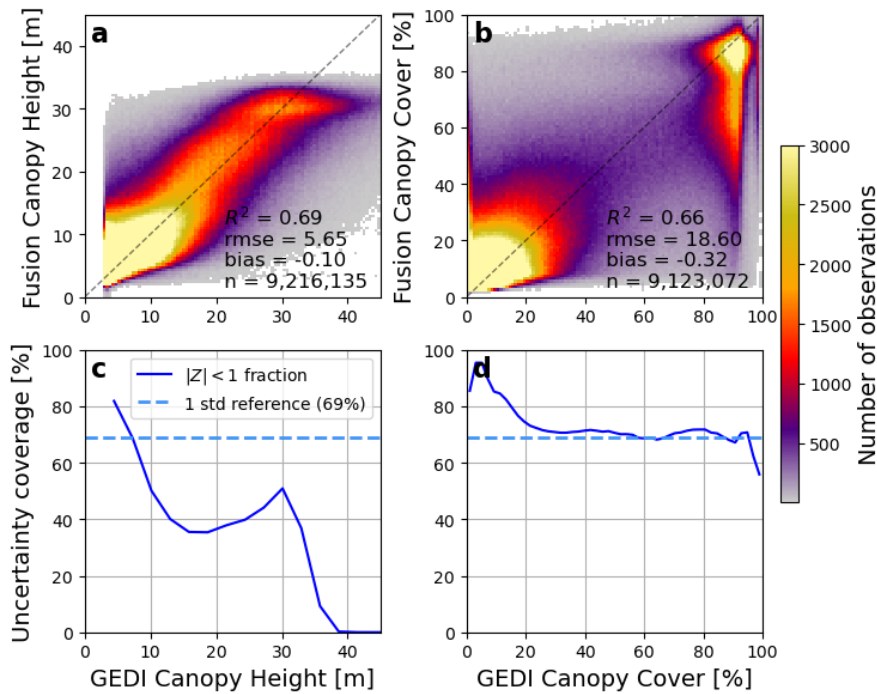

Figure 13. Transfer learning performance for GEDI canopy height (RH98) and canopy cover predictions using efficient transfer where feature extraction layers were frozen and only the regression head of the global WSCI model was updated. Panels show: (a,b) relationship between predicted and observed height and cover values; and (c,d) uncertainty coverage across the range of canopy height and cover values.

## 4. Discussion

This study demonstrates that multimodal deep learning can effectively map forest structural complexity at high spatial resolution on a global scale by fusing sparse GEDI lidar data with wall-to-wall SAR imagery. Our approach balances high predictive performance with model interpretability and efficiency, integrating predictive uncertainty, input explainability, and a compact architecture of fewer than 400,000 parameters. This efficiency, enabled by the EfficientNetV2 backbone (Figure 2) and training on more than 130 million GEDI footprints, allows accurate predictions across diverse regions and time periods without requiring high-end computing resources.

Previous studies that employed deep learning techniques to upscale GEDI variables wall-to-wall primarily focused on canopy height [14,21] and biomass [15,55], achieving $R^2$ values between 0.5 and 0.8. Global models often rely on single-mode inputs and deeper network architectures than our EfficientNetV2 [14,15]. Conversely, regional multimodal approaches have conducted minimal or no assessments of model performance over time [56–58]. Our work leverages multimodal datasets at a global scale, coupled with extensive temporal and geographical validation by assessing model performance across biomes over multiple years using both GEDI and ALS data for mapping forest structural complexity. Moreover, our transfer learning tests adapting the global model pre-trained on WSCI to predict canopy height and canopy cover achieved $R^2$ values of 0.69 and 0.66 (Figure 13), respectively, with substantially faster training and fewer training epochs to achieve model convergence compared to models trained from random weight initialization, with further gains in computational performance by retraining only the weights in the model's regression head. These results suggest that a large and geographic diverse training dataset, combined with multimodal inputs, may be more important in achieving consistently accurate estimates than the specific architecture or depth of the network. Given its relatively compact structure and modest computational demands, the EfficientNetV2 model may be well suited for integration into forest monitoring systems, particularly where resources are limited.

Large-scale products that interpolate sparse GEDI observations often fail to capture local spatial variation and forest-to-non-forest transitions [59], while temporal monitoring has been limited across areas underrepresented by GEDI data (e.g. boreal forests) or affected by fine-scale disturbances [60,61]. Our model overcomes these spatial limitations by providing continuous 25-meter resolution estimates with global coverage, enabling detailed analyses of forest structure in heterogeneous landscapes while capturing local spatial patterns (Figure 11) and forest-to-non-forest transitions (Figure 12, Sup. Fig. 6). To address temporal monitoring gaps, our model's design balances multimodal inputs by leveraging the complementary strengths of different SAR products, with our feature importance analysis underscoring the critical role of high temporal resolution data (Sentinel-1) that enable change assessment in sub-annual intervals that which can be used to support temporal monitoring of post-disturbance dynamics (Sup. Fig. 9). Furthermore, we incorporate geographical priors as model inputs to regularize model predictions across global regions [14]. By providing wall-to-wall estimates of WSCI at high temporal resolution, we facilitate ecological studies to map local gradients of structural complexity anywhere, as well as distinguish between phenological variations and changes in canopy structure, supporting both short-term monitoring of disturbance effects within annual cycles and consistent long-term trend assessments across multiple years.

The ability to extract pixel-by-pixel time series of WSCI is a significant advance (Figure 6 and Sup. Fig. 9). This continuous monitoring capability may provide insight into how canopy structure changes over time and help understand the causes and implications of these dynamics. In this sense, such a capability complements existing products primarily focused on monitoring gain/loss of forest area [62,63]. This observational record may help inform policy and management by highlighting regions that are either stable hotspots of high complexity, and thus should be protected, or areas undergoing early degradation, thereby enabling proactive intervention through management practices that prevent further structural losses and promote forest recovery.

Forest structural complexity cannot be directly observed at scale, but as illustrated here, it can be inferred from sensors sensitive to vertical structure. Our approach uses lidar and SAR to estimate structural complexity globally, though several enhancements could further improve model fidelity and generalizability. First, incorporating optical data from Landsat and Sentinel-2 could address phenological variations in canopy foliage, as these datasets have successfully modeled forest structural variables in previous studies [14,64], albeit with potential computational efficiency tradeoffs. Second, our current model relies solely on SAR backscatter information; integrating additional SAR-derived information such as interferometric coherence and temporal decorrelation (e.g., [65]) could enhance prediction accuracy and robustness. Finally, upcoming spaceborne missions offer promising opportunities to address current limitations: NISAR will provide L-band measurements globally at high temporal resolution [66] for continuous spatiotemporal estimates, while BIOMASS P-band measurements (~70cm) [67] will likely overcome signal saturation issues in high biomass areas that currently limit SAR-based products.

Our study advances high-resolution, large-scale mapping of forest structure from space through deep learning, providing comprehensive insights into four key aspects: the geographical and temporal generalizability of model predictions; the relative contribution of multiple information sources; the ability to capture local spatial gradients from a global model; and the potential for transfer learning to map multiple forest attributes. Despite these advancements, several limitations constrain our findings and point toward important areas for future development. First, regarding temporal generalizability, not all input layers share the same temporal resolution, as sub-annual mosaics were only available through Sentinel-1 data in Google Earth Engine. This temporal mismatch makes it unclear whether observed seasonal variations truly reflect forest phenology or are artifacts of signal saturation [21,68]. Second, our assessment of geographical generalizability was limited in boreal regions where GEDI coverage is sparse. Further validation with ALS datasets collected across different seasons in these regions is needed, and incorporating additional data sources such as ICESat-2 [69] may improve model reliability in these undersampled zones. Third, while our model successfully integrates multiple information sources, the propagation of uncertainties through sequential modeling stages remains challenging. Although the GEDI L4C product provides uncertainty estimates at the footprint level [29], combining these with uncertainties from deep learning models introduces additional complexity requiring methodological development. Future work should incorporate techniques for leveraging multiple uncertainty sources, including bootstrapping methods that explicitly incorporate input uncertainty during training [70] and regularization approaches that use known input uncertainties to calibrate deep neural network estimates [71]. These methods would enable more robust uncertainty propagation from initial GEDI observations through multiple modeling stages.

## 5. Conclusions

This study demonstrates the potential of multimodal deep learning for global forest monitoring by fusing GEDI observations with SAR datasets to address limitations in current forest structural complexity mapping. We developed a scalable, computationally efficient model with explicit uncertainty quantification that extends coverage beyond existing GEDI observations while maintaining reasonable accuracy with minimal computational overhead. The fusion approach underscores the efficacy of combining complementary SAR wavelengths, with L-band PALSAR providing deep canopy penetration and C-band Sentinel-1 offering high temporal resolution, creating synergies that individual sensors cannot achieve. The framework's transferability suggests potential for predicting multiple forest attributes and provides a methodological foundation for integrating data from upcoming missions such as NISAR and BIOMASS, which may advance comprehensive forest monitoring capabilities. The resulting wall-to-wall, multi-temporal datasets address gaps in existing monitoring frameworks by providing spatial and temporal detail that could support ecological applications ranging from biodiversity conservation to ecosystem management under climate change, potentially enabling forest structural dynamics to be monitored continuously at management-relevant scales.

## Acknowledgements

We gratefully acknowledge the funding from National Aeronautics and Space Administration (NASA) contract NNL 15AA03C for the development and execution of the GEDI mission, including funding to R.D. and J.A., and NASA FINNEST grant 80NSSC22K1543 to T.C.

# Supplementary information

Supplementary Table 1. Performance metrics of the GEDI-SAR WSCI fusion model across major biomes. For each biome, we report summary statistics (mean, standard deviation, minimum, and maximum) for three key metrics: coefficient of determination ($R^2$), root mean square error (RMSE), and uncertainty coverage (|Z-score| < 1 fraction). Statistics were calculated from quarterly validation periods between April 2019 and December 2022. The rightmost column (N) indicates the number of validation samples within each biome. Performance varied across biomes, with tropical moist forests showing the highest mean $R^2$ (0.75) and lowest RMSE (0.47), while Tundra regions exhibited substantially lower performance, likely due to limited training data and sparse vegetation structure.

| Biome | $R^2$ | | | | RMSE | | | | |Z-score| fraction < 1 | | | | N |
|---|---|---|---|---|---|---|---|---|---|---|---|---|---|
| | mean | std | min | max | mean | std | min | max | mean | std | min | max | |
| Tropical & Subtropical Moist Broadleaf Forests | 0.75 | 0.03 | 0.72 | 0.80 | 0.47 | 0.03 | 0.42 | 0.51 | 0.74 | 0.01 | 0.72 | 0.76 | 5,742,359 |
| Tropical & Subtropical Dry Broadleaf Forests | 0.64 | 0.05 | 0.54 | 0.70 | 0.58 | 0.03 | 0.54 | 0.64 | 0.69 | 0.01 | 0.67 | 0.72 | 849,751 |
| Tropical & Subtropical Coniferous Forests | 0.66 | 0.03 | 0.59 | 0.72 | 0.57 | 0.03 | 0.51 | 0.67 | 0.71 | 0.03 | 0.65 | 0.79 | 219,266 |
| Temperate Broadleaf & Mixed Forests | 0.67 | 0.04 | 0.60 | 0.75 | 0.60 | 0.02 | 0.57 | 0.63 | 0.72 | 0.01 | 0.70 | 0.74 | 3,493,413 |
| Temperate Conifer Forests | 0.68 | 0.02 | 0.65 | 0.72 | 0.60 | 0.01 | 0.58 | 0.62 | 0.70 | 0.01 | 0.68 | 0.72 | 1,549,736 |
| Boreal Forests/Taiga | 0.59 | 0.06 | 0.47 | 0.72 | 0.54 | 0.03 | 0.49 | 0.58 | 0.67 | 0.02 | 0.64 | 0.72 | 983,207 |
| Tropical & Subtropical Grasslands, Savannas & Shrublands | 0.69 | 0.04 | 0.63 | 0.76 | 0.48 | 0.01 | 0.45 | 0.50 | 0.68 | 0.02 | 0.66 | 0.71 | 4,839,715 |
| Temperate Grasslands, Savannas & Shrublands | 0.62 | 0.05 | 0.55 | 0.68 | 0.47 | 0.04 | 0.41 | 0.54 | 0.70 | 0.02 | 0.68 | 0.73 | 2,431,326 |
| Flooded Grasslands & Savannas | 0.63 | 0.08 | 0.52 | 0.77 | 0.49 | 0.03 | 0.45 | 0.54 | 0.71 | 0.03 | 0.66 | 0.75 | 239,162 |

| | | | | | | | | | | | | |
|---|---|---|---|---|---|---|---|---|---|---|---|---|
| Montane Grasslands & Shrublands | 0.60 | 0.03 | 0.55 | 0.65 | 0.48 | 0.02 | 0.44 | 0.51 | 0.69 | 0.01 | 0.66 | 0.72 | 1,327,441 |
| Tundra | 0.21 | 0.11 | 0.00 | 0.40 | 0.69 | 0.19 | 0.45 | 1.20 | 0.67 | 0.12 | 0.38 | 0.83 | 15,588 |
| Mediterranean Forests, Woodlands & Scrub | 0.61 | 0.04 | 0.55 | 0.68 | 0.54 | 0.02 | 0.52 | 0.57 | 0.69 | 0.01 | 0.67 | 0.71 | 1,247,366 |
| Deserts & Xeric Shrublands | 0.48 | 0.05 | 0.38 | 0.55 | 0.44 | 0.04 | 0.38 | 0.51 | 0.68 | 0.02 | 0.66 | 0.72 | 3,620,769 |
| Mangroves | 0.69 | 0.05 | 0.59 | 0.77 | 0.55 | 0.03 | 0.48 | 0.60 | 0.73 | 0.03 | 0.69 | 0.78 | 187,822 |

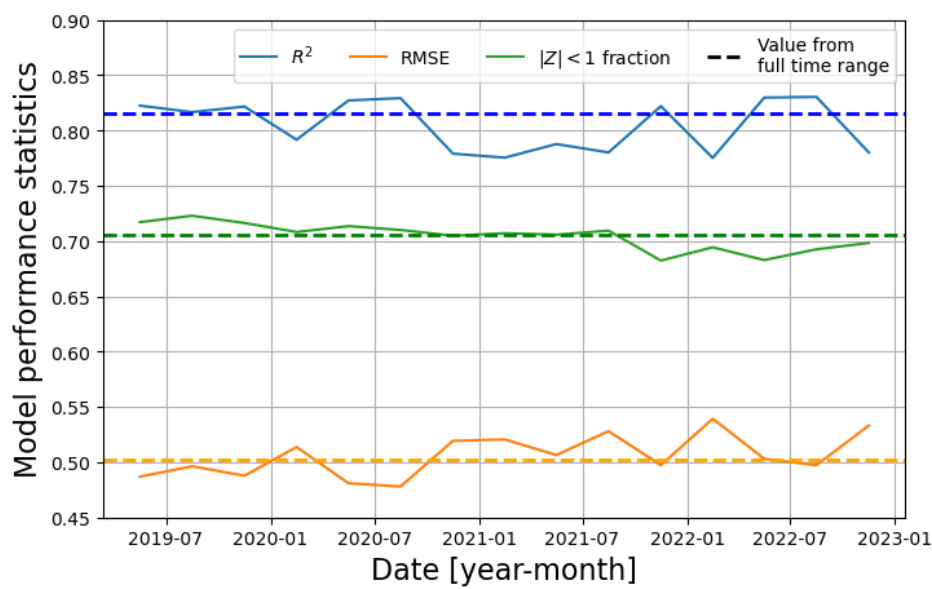

Supplementary Figure 1. Temporal stability of the GEDI-SAR WSCI fusion model performance across quarterly periods from 2019-2023. The plot shows three key performance metrics: coefficient of determination ($R^2$) in blue, root mean square error (RMSE) in orange, and uncertainty coverage (|Z-score| < 1 fraction) in green. Dashed horizontal lines indicate the average value for each metric across the entire time range of GEDI data used in this study. Despite minor fluctuations, all metrics remain consistent throughout the study period, demonstrating the model's robust temporal transferability and minimal seasonal bias.

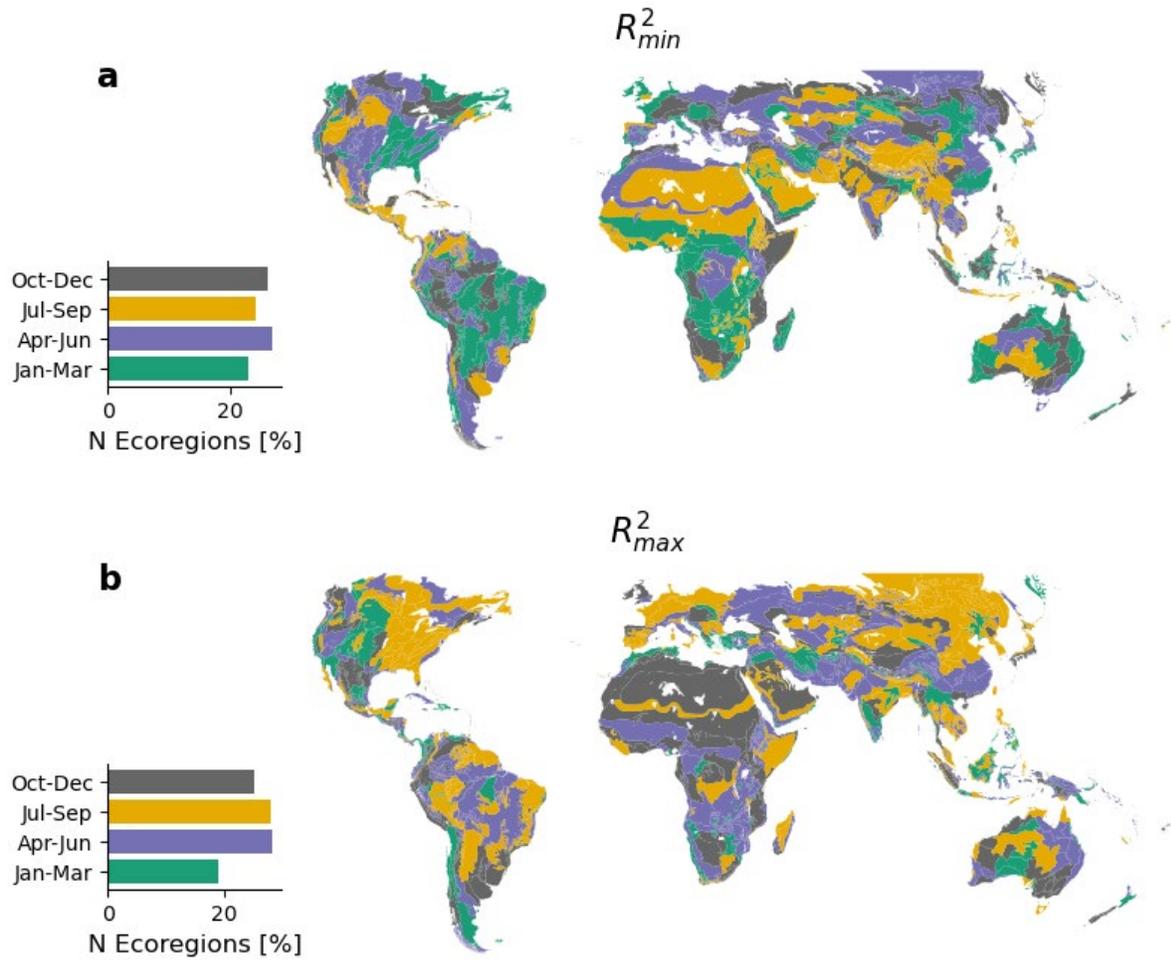

Supplementary Figure 2. Seasonal variations in model performance across biomes within GEDI's orbital coverage (51.6°N to 51.6°S). Maps show (a) minimum and (b) maximum coefficient of determination ($R^2$) values observed for each ecoregion across quarterly periods from 2019-2023, colored by the season in which these extremes occurred. The bar charts show that optimal model performance (maximum $R^2$) occurred most frequently during the April-September period.

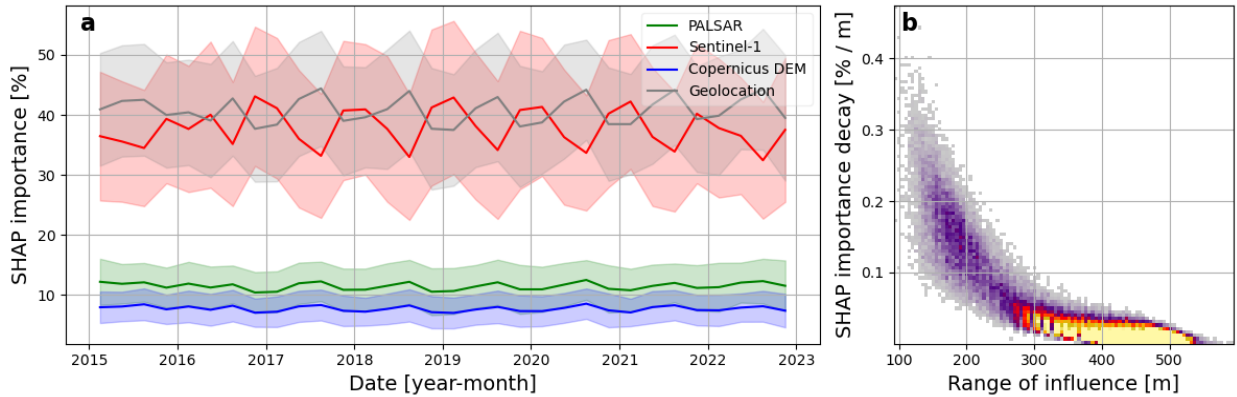

Supplementary Figure 3. Seasonal SHAP importance patterns at (a) high latitudes (>52°N), and (b) SHAP importance from spatial features within 1×1 km prediction windows. (SHAP importance decay = decrease in influence of an input pixel over a prediction as their distance increases; Range of influence = distance where feature importance reaches its minimum).

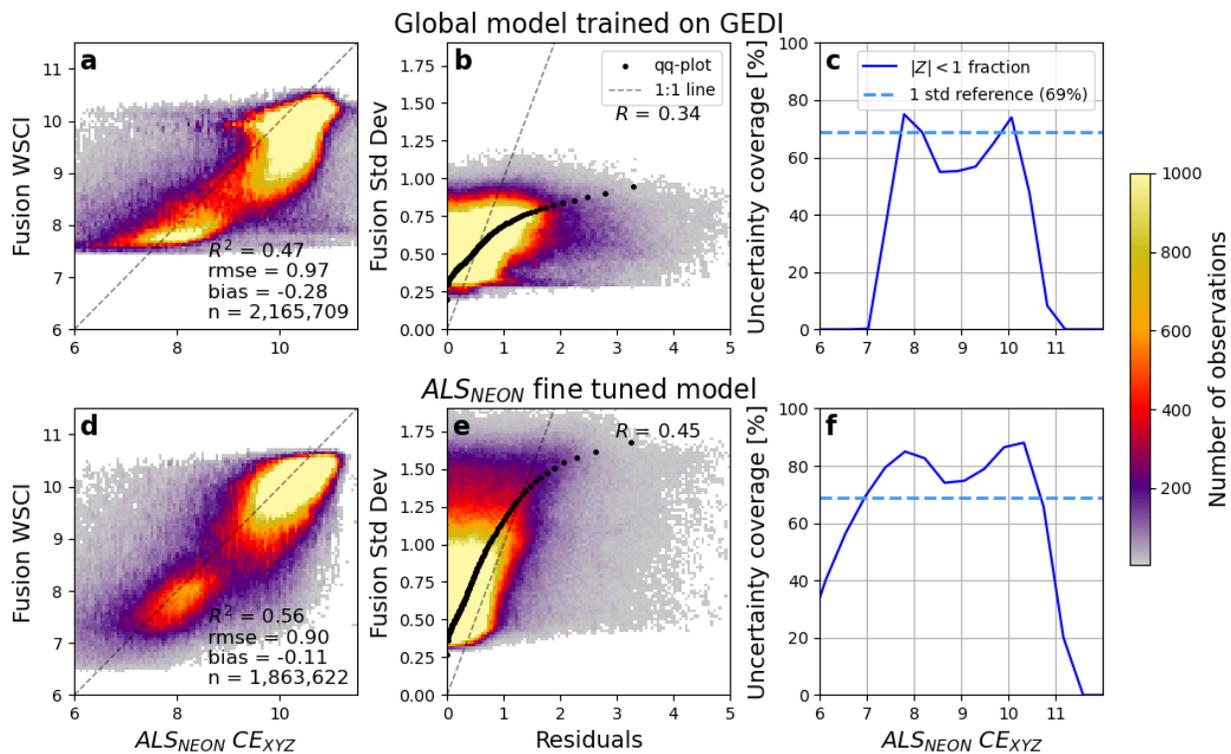

Supplementary Figure 4. Model performance of the GEDI-SAR WSCI Fusion against structural complexity measurements ($CE_{XYZ}$) from ALS data in the NEON network over the United States (a-c) before and (d-f) after fine tuning.

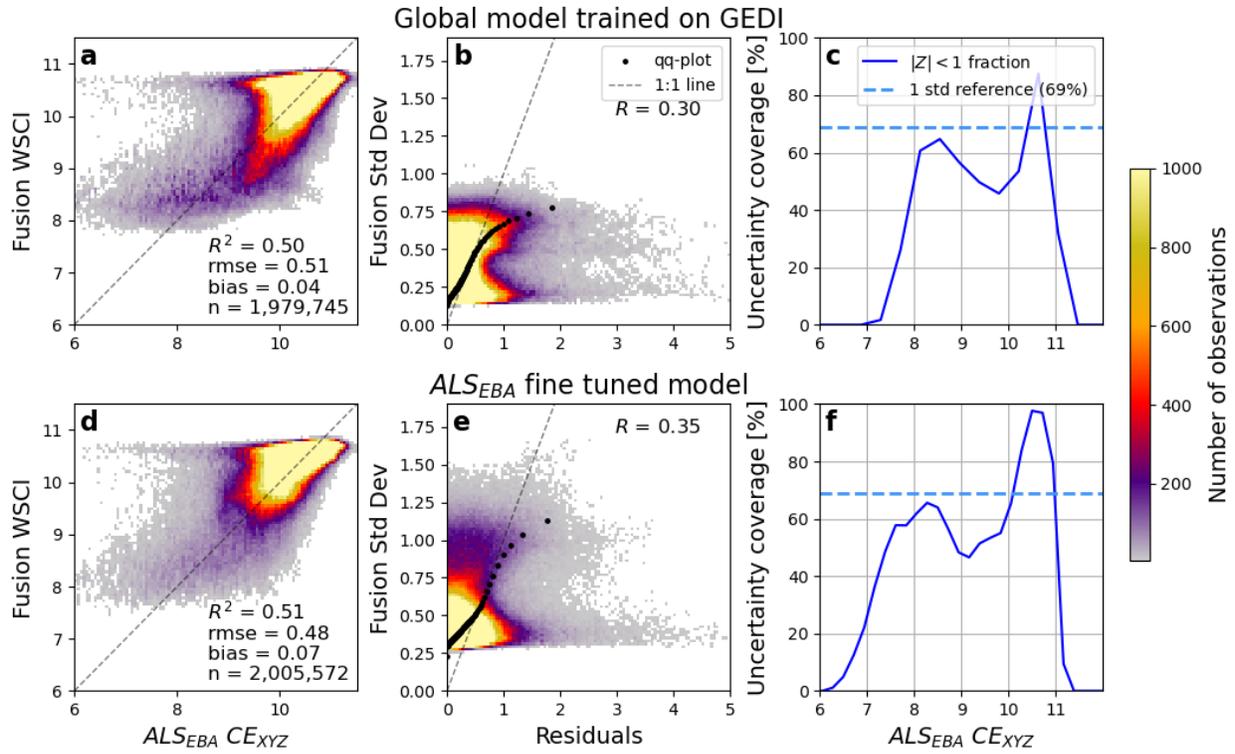

Supplementary Figure 5. Model performance of the GEDI-SAR WSCI Fusion against structural complexity measurements ($CE_{XYZ}$) from ALS data in the EBA network over the Brazilian Amazon rainforest (a-c) before and (d-f) after fine tuning.

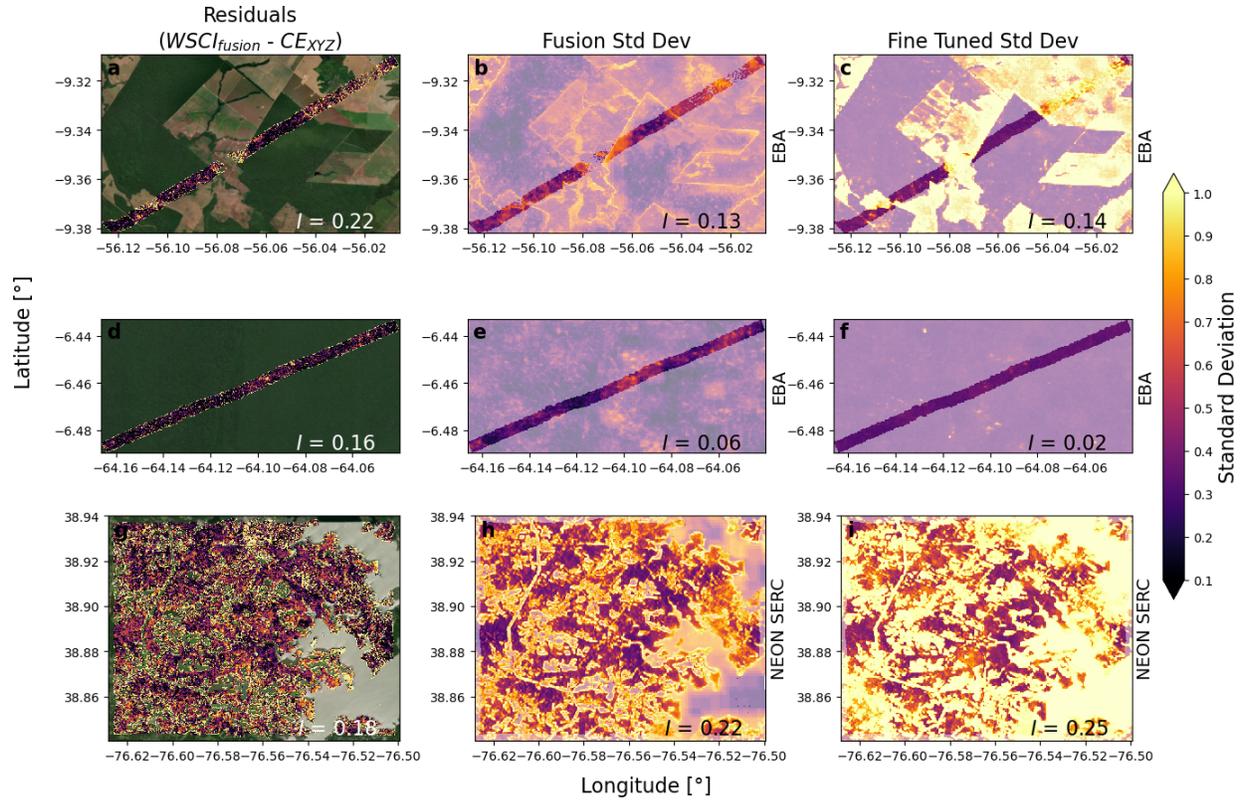

Supplementary Figure 6. Wall-to-wall uncertainty patterns at 25 m resolution for the same sites as Figure 15. Columns show: (a,d,g) absolute residuals between fusion estimates and ALS measurements, (b,e,h) predicted standard deviation from the global fusion model, and (c,f,i) predicted standard deviation after regional fine-tuning. Spatial correlation (I) values in the left column (a,d,g) represent the residuals autocorrelation, and indicate the alignment between predicted uncertainty and model residuals in the center and right columns.

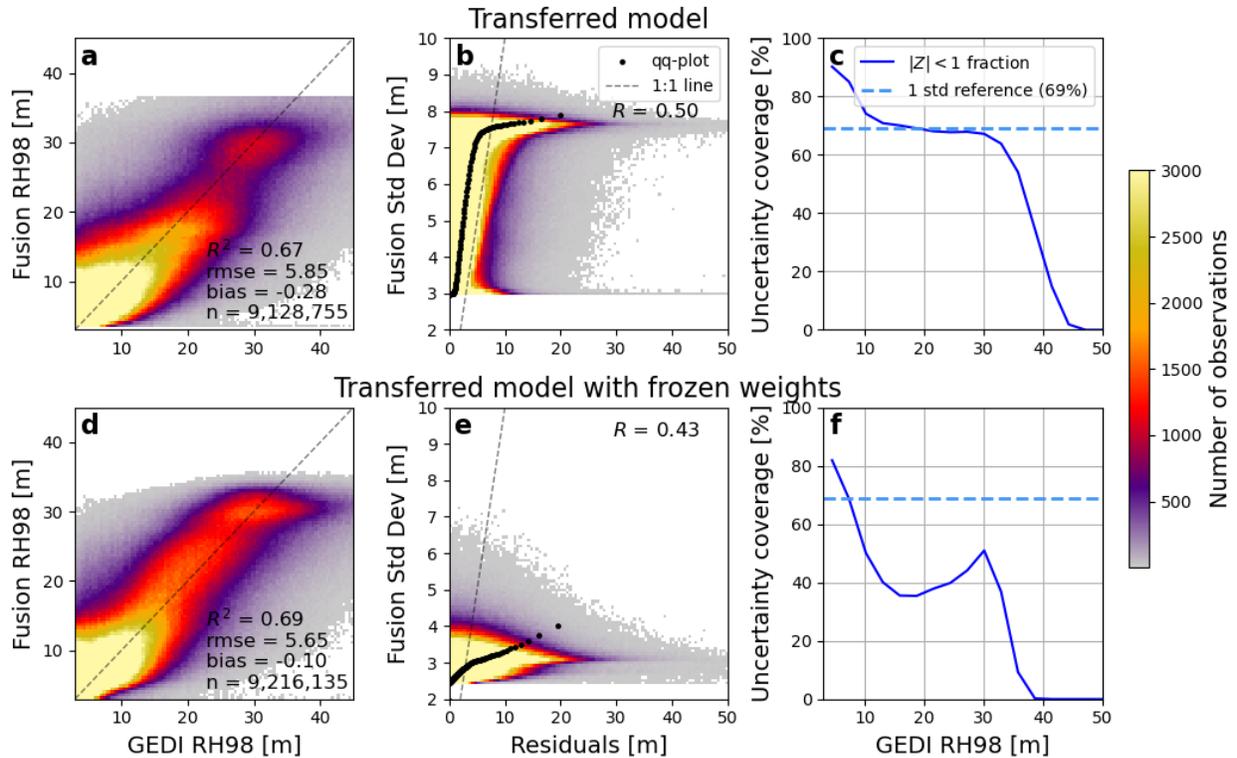

Supplementary Figure 7. Transfer learning performance for GEDI canopy height (RH98) prediction. Comparison between: (a-c) full-model transfer where all network weights were updated during training, and (d-f) efficient transfer where feature extraction layers were frozen and only the regression head was updated. For each approach, panels show: (a,d) relationship between predicted and observed height values; (b,e) uncertainty calibration between predicted standard deviation and actual residuals; and (c,f) uncertainty coverage across the range of canopy heights. Both approaches achieve similar explanatory power ($R^2 \approx 0.67$-$0.69$) but with the frozen-weights model showing slightly lower bias and the fully transferred model showing better calibrated uncertainty estimates.

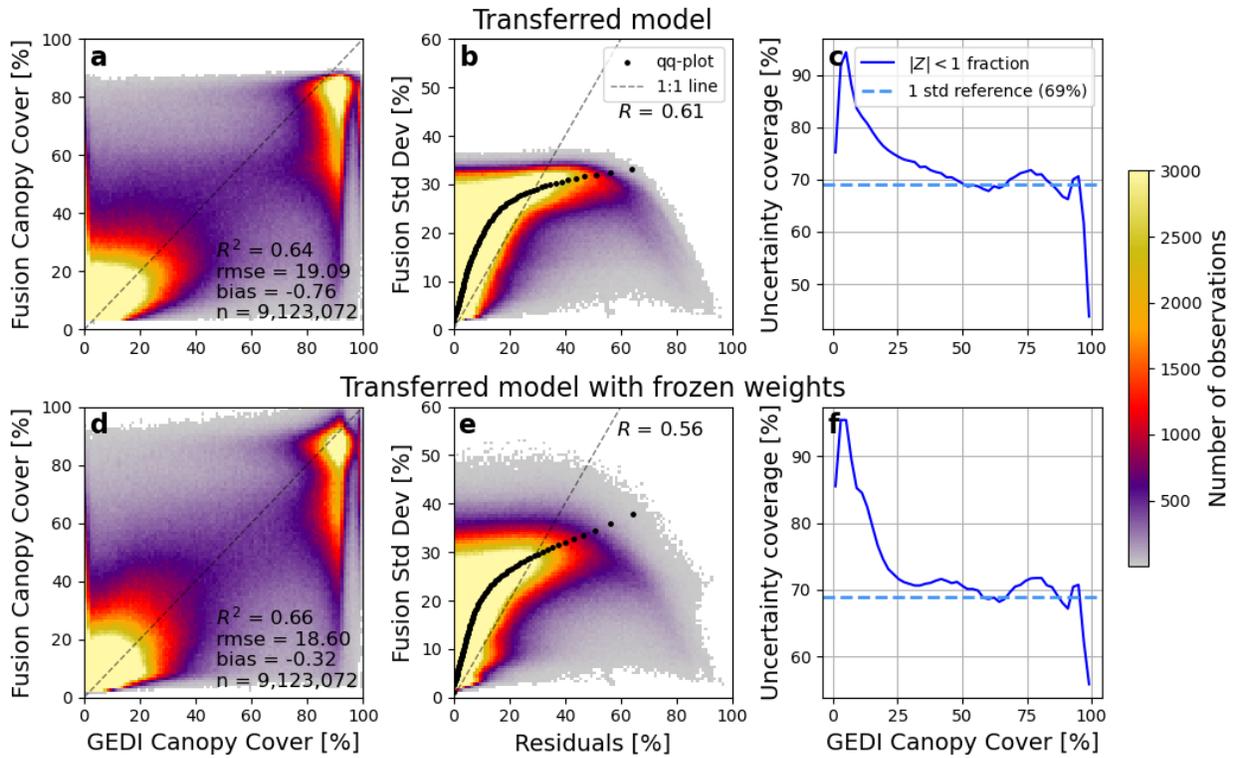

Supplementary Figure 8. Transfer learning performance for GEDI canopy cover prediction. Comparison between: (a-c) full-model transfer with complete weight updates, and (d-f) efficient transfer with frozen feature extraction layers. For each approach, panels show: (a,d) relationship between predicted and observed canopy cover; (b,e) uncertainty calibration; and (c,f) uncertainty coverage across cover percentages. Both approaches achieve similar explanatory power ($R^2 \approx 0.64$-0.66), with the frozen-weights model showing slightly better performance despite requiring substantially less computational effort.

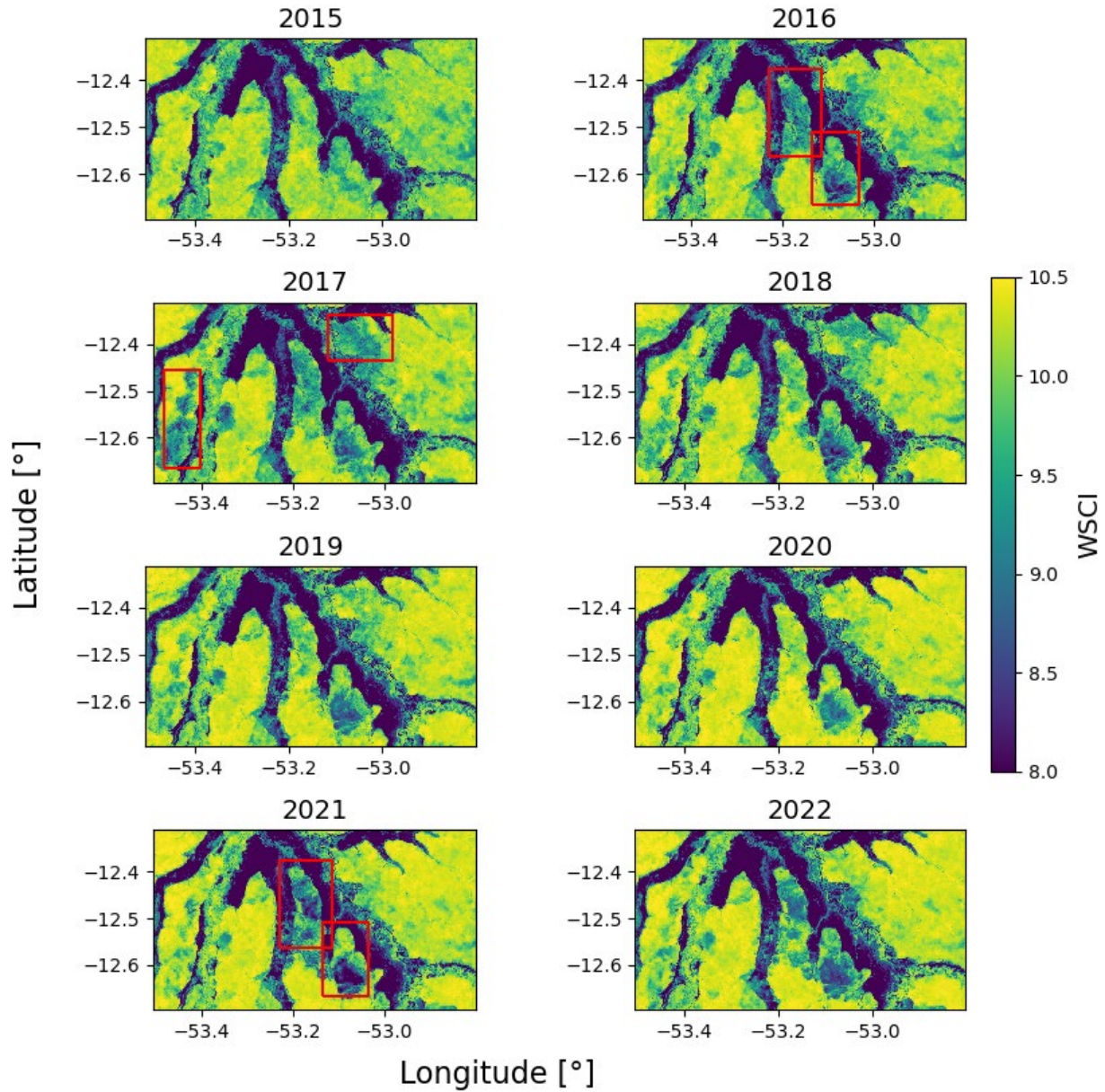

Supplementary Figure 9. Time series of forest structural complexity in the Xingu Indigenous Territory, Brazilian Amazon (2015-2022). Each panel represents the WSCI values for a specific year at 25m resolution, revealing both stable forest areas and dynamic changes in structural complexity. Red rectangles highlight notable disturbance events from recurring forest fires in 2016, 2017 and 2021, causing significant reductions in structural complexity. The ability to generate such continuous time series at high spatial resolution enables precise monitoring of forest structural dynamics, including disturbance events, recovery patterns, and gradual changes that would be difficult to detect with discrete observations.